\documentclass[onecolumn]{fairmeta}

\usepackage[most]{tcolorbox}

\usepackage{amsmath}
\usepackage{amssymb}
\usepackage[utf8]{inputenc} 
\usepackage[T1]{fontenc}    
\usepackage{hyperref}       
\usepackage{url}            
\usepackage{booktabs}       
\usepackage{amsfonts}       
\usepackage{nicefrac}       
\usepackage{microtype}      
\usepackage{xcolor}         
\usepackage{graphicx}
\newcommand{\p}[1]{}

\usepackage{wrapfig}
\usepackage{enumitem}
\usepackage{fontawesome5}

\usepackage{amsthm}   

\theoremstyle{remark}

\usepackage{algorithm}
\usepackage{algpseudocode}   

\renewcommand{\algorithmiccomment}[1]{\textbf{Initialize:}}  

\title{\LARGE Physics Filtering Favors the Generalization of Robot Learning}  

\author[1,2,*,\ddag]{\href{https://jiajindou.github.io/}{\textcolor{black}{Jindou Jia}}}
\author[2,*]{Shixuan Han}
\author[2,*]{Meng Wang}
\author[1]{\href{https://reagan1311.github.io/}{\textcolor{black}{Gen Li}}}
\author[2]{Zihan Yang}
\author[3]{Sicheng Zhou}
\author[2,\dagger]{Kexin Guo}
\author[1,\dagger]{Jianfei Yang}
\author[2]{Xiang Yu}
\author[4]{Wei Wang}
\author[2]{Lei Guo}

\affiliation[1]{Nanyang Technological University}
\affiliation[2]{Beihang University}
\affiliation[3]{National University of Singapore}
\affiliation[4]{Beijing Aerospace Control Instrument Research Institute}

\contribution[*]{Equal Contribution}
\contribution[\ddag]{Project Lead}

\metadata[GitHub]{\url{https://github.com/JIAjindou/PhyFilter}}
\metadata[Webpage]{\url{https://scoardyy.github.io/PhyFilter}}

\correspondence{Jianfei Yang, \email{jianfei.yang@ntu.edu.sg}; Kexin Guo, \email{kxguo@buaa.edu.cn}.}

\abstract{
  Living organisms exhibit extraordinary adaptability to unseen environments through their intrinsic physical structures and lifelong feedback-driven learning. Endowing robots with comparable generalization is critical for reliable operation in the real world. While recent approaches attempt to improve generalization by scaling training data, such strategies remain impractical for robotics, where collecting real-world demonstrations at the scale of large language models is prohibitively costly and slow. Contrary to this reliance on massive datasets, we show that robots can generalize effectively under dynamics uncertainties even with limited training data by leveraging a feedback mechanism, namely PhyFilter, that corrects learning outputs with physics-filtered learning residuals. PhyFilter operates as a lightweight, model-agnostic module whose parameters can be automatically optimized through an auto-learning algorithm, eliminating manual tuning and enabling seamless integration with diverse robot policies. We validate PhyFilter across four representative robotic systems, demonstrating that it enables quadruped robots to generalize to unseen terrains, payload variations, and speed ranges; drones to flight under unseen wind disturbances; aerial manipulators to achieve centimeter-level in-air capture despite wind and mass uncertainties; and acceleration differentiators to remain robust with distribution shift. These results show that physics-filtered feedback can serve as a powerful alternative to massive data scaling.
}

\begin{document}

\maketitle

\section{Introduction}\label{sec_intro}


Over the past decade, large language models (LLMs)~\citep{achiam2023gpt, team2023gemini, guo2025deepseek} have attained remarkable advancements, primarily fueled by extensive training datasets, large-scale neural networks, and massive computational power~\citep{bahri2024explaining, lecun2015deep, wang2023scientific}. As robots serve as the physical instantiation of intelligence, delineating a feasible pathway toward human-level robotic intelligence has emerged as a pivotal and intuitive goal~\citep{kaufmann2023champion, choi2023learning, radosavovic2024real, jia2026action}. A core prerequisite for such machine intelligence lies in generalization, robust adaptation to unseen environments, a critical capability for reliable real-world deployment~\citep{zador2022toward}. With network architectures and graphics processing unit hardware now relatively mature, a fundamental question naturally arises: Can scaling data alone enable generalization in robotics?


Unlike the text corpora used to train LLMs, which are abundant and readily available on the internet, robot learning requires massive training data collected in the physical world. Currently, robot data scaling efforts follow two primary paths. The first is to collect data through human teleoperation~\citep{zhao2023learning, lin2024data}, but obtaining human demonstrations at LLM scale (billions to hundreds of billions of tokens) remains technically difficult, time-consuming, and costly~\citep{kaelbling2020foundation, dreczkowski2025learning}. Even large industrial efforts, such as Tesla’s teleoperation-assisted factory data pipeline, face fundamental bottlenecks in throughput and scalability~\citep{tesla_2025}. The second path relies on synthetic data from simulation engines (e.g., Isaac Lab, MuJoCo)~\citep{todorov2012mujoco, makoviychuk2021isaac, geng2025roboverse} and video-based generation methods (e.g., OpenAI Sora, Cosmos world model)~\citep{opensora2, nvidia2025cosmosworldfoundationmodel, jain2024vid2robot, chen2025vidbot, hou2026world,WM_Survey}. Although these approaches yield performance improvements on certain robotic tasks~\citep{tobin2018domain, chen2025vidbot, ai2025review}, they remain limited in scaling robot generalization due to the persistent sim-to-real gap, prohibitive computational overhead, and the lack of standardized data formats, software, and hardware interfaces across diverse robotic platforms~\citep{zador2022toward}.

From a bionic point of view, the powerful generalization capabilities of living organisms do not rely solely on the scaling law~\citep{zador2022toward, meister2022learning, SOMOGYI2015186}. It has been observed that innate or acquired physical structure emerges to facilitate the adaptation of living organisms to unseen environments~\citep{thuerey2021physics,  heeger2017theory, mejias2016feedforward, markov2021cerebellar}. For example, cortical neural responses are governed by the synergistic integration of feedforward signals, feedback loops, and prior drives, which evolve dynamically to minimize a global energy function~\citep{heeger2017theory}. Likewise, larval zebrafish can integrate their innate physical integral structure and real-time state feedback with an updated cerebellar internal model to effectively respond to unpredictable visual perturbations~\citep{markov2021cerebellar}. Inspired by these insights, intelligent robots should also utilize readily accessible physics structures or other priors~\citep{dreczkowski2025learning, kaelbling2020foundation, marshall2025transformers}, instead of depending exclusively on data scaling. Different from digital LLMs, real-world robots indeed possess distinct, well-established physical structures and real-time feedback resulting from environmental interactions that can be leveraged. This leads to an interesting question: How to harness these physical architectures to improve robots' learning generalization?

\begin{figure*}
	\centering
	\includegraphics[width=1\linewidth]{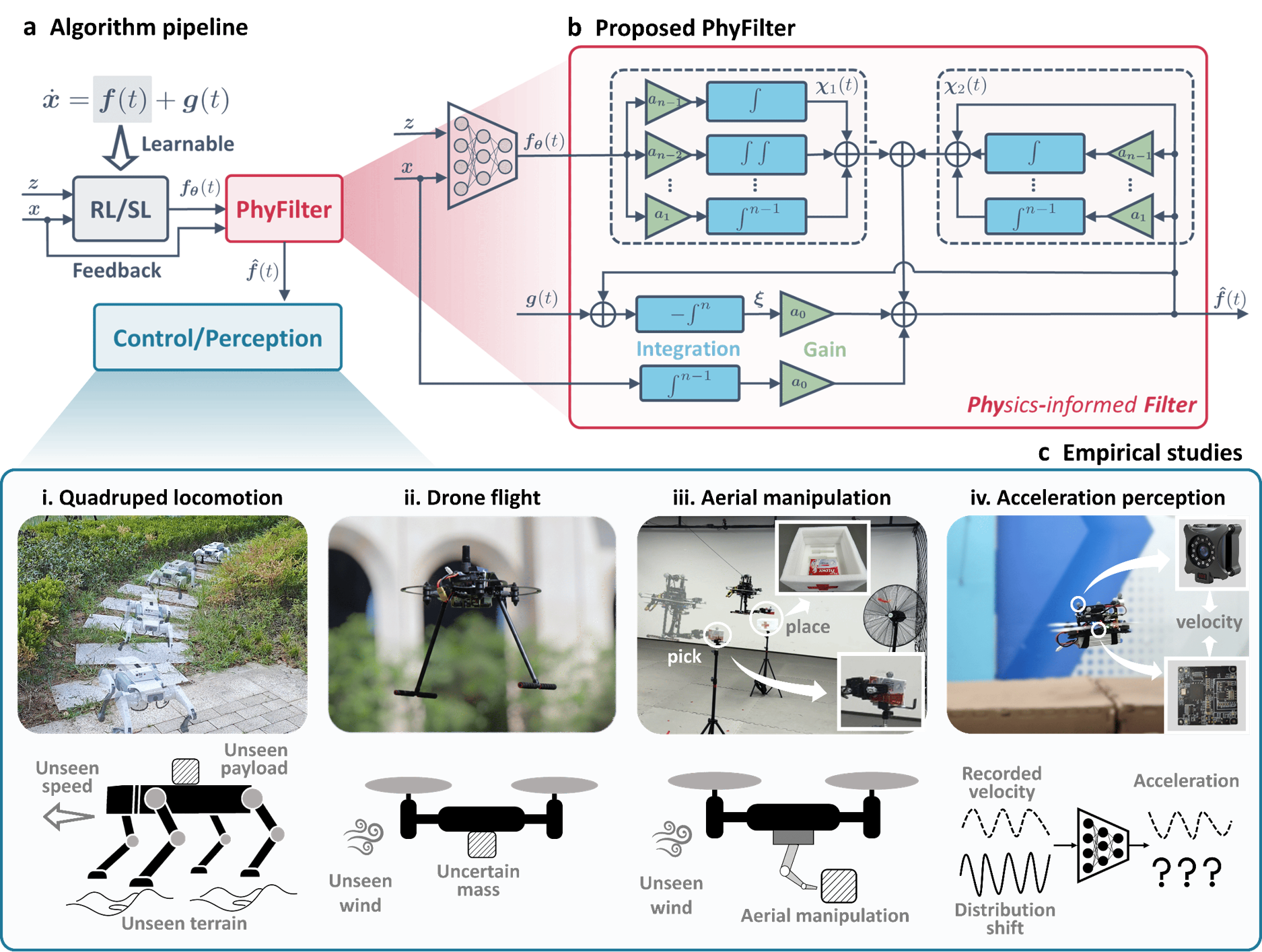}
	\caption{\textbf{Overview of our PhyFilter.} \textbf{a}, The learning output of RL or SL is enhanced by a physics-informed filter, which helps narrow the large generalization gap. \textbf{b}, The implementation details of PhyFilter, which is analytic and convergent. Additional theoretical details are provided in the ``Filtering learning residual'' section in Methods. \textbf{c}, PhyFilter is validated across four representative examples. \textbf{i}, Policy learning for quadruped locomotion. A quadruped robot trained in a simplified simulator generalizes to unseen speeds, payload variations, and real-world terrains when equipped with PhyFilter. \textbf{ii}, Dynamical learning for drone maneuvering flight. A learning-based model captures mass variations, while PhyFilter helps handle unseen wind disturbance. \textbf{iii}, Dynamical learning for aerial manipulation in a pick-and-place task. A dedicated model learns drone-manipulator coupling uncertainties, with PhyFilter aiding in the handling of unseen wind disturbances and mass uncertainties. \textbf{iv}, Kinematics learning for acceleration perception. A neural network differentiator maps velocity sequences to real-time acceleration, and PhyFilter improves its robustness to distribution shift. \textcolor{black}{We also demonstrate the performance of PhyFilter on a humanoid robot initially, which is detailed in Supplementary Fig. S6.}}
	\label{All_scenarios}
\end{figure*}

\textcolor{black}{We propose a novel \textbf{Phy}sics \textbf{Filter}ed learning approach, termed PhyFilter, which enhances both generalization under dynamics uncertainties and interpretability of robot learning. Concretely, PhyFilter operates by correcting learning outcomes according to readily accessible physical differential structure and real-time state feedback~\citep{NDO_constant, DOBC_survey},} as shown in Fig. \ref{All_scenarios} (a and b). By framing the method within a filtering-like paradigm, PhyFilter enhances interpretability and improves convergence. We also present an auto-learning algorithm capable of searching for the optimal parameters of PhyFilter, making manual tuning unnecessary. Unlike prior physics-informed methods~\citep{wang2024imperative, greydanus2019hamiltonian, cranmer2020lagrangian, romero2024actor} that typically require a retraining of the modified network, PhyFilter features a plug-and-play design, enabling direct deployment with pretrained models.



PhyFilter applies seamlessly to two prevailing learning paradigms, reinforcement learning (RL) and supervised learning (SL), and has been demonstrated to be effective in four representative robotic scenarios. Experiments include quadruped locomotion~\citep{han2024lifelike, yang2020multi, kumar2021rma, lee2024learning, kim2025high, shi2024rethinking, rudin2022learning}, drone maneuvering flight~\citep{jia2025, 9779449}, aerial manipulation~\citep{9462539}, and acceleration perception~\citep{9635983}, as depicted in Fig. \ref{All_scenarios}c. Specifically, with the assistance of PhyFilter, a policy trained solely on simulated flat terrain can stably generalize to diverse real-world terrains on a quadruped robot. Moreover, an aerial manipulator can perform precise pick-and-place tasks with a maximum manipulation error of $2.5$ cm, a $23.99\%$ \textit{avg.} and $55.04\%$ \textit{var.} improvement over baseline, despite $5$ m/s wind and $0.3$~kg mass uncertainties. Experimental videos and details are provided in Supplementary Information.

\section{Results}\label{sec_result}

\subsection{{{Phy}sics {filter}ed learning}}

We begin by introducing the core concept of PhyFilter. Consider an unknown mapping $\bm{f}(t)$ appearing in governing physical equations. The learning residual $\bm{\gamma}(t)$ in unseen scenarios can be formalized as $\bm{\gamma}(t) = \bm{f}(t) - \bm{f}_{\bm{\theta}}(t)$, with the truth model $\bm{f}(t)$ and learned model $\bm{f}_{\bm{\theta}}(t)$ parameterized by parameter $\bm{\theta}$. \textcolor{black}{$\bm{\gamma}(t)$ arises from unseen factors such as sensor noise, external disturbances, or unmodeled interactions.}

Intuitively, if real-time knowledge of $\bm{\gamma}(t)$ is accessible, the model output could be corrected online as $\bm{f}_{\bm{\theta}}(t) + \bm{\gamma}(t)$. In practice, however, the absence of labels during deployment makes this correction infeasible, as $\bm{\gamma}(t)$ cannot be directly inferred. Even natural organisms are unable to instantly interpret unfamiliar environments, but instead undergo an adaptation process. Previous studies have discovered a low-pass filtering feature in human perception systems, such as the vision~\citep{braje1995human} and hearing~\citep{peterson2020phase}. These observations suggest that filtering may underlie the adaptive mechanisms enabling humans to remain robust under environmental variations. In this work, we try to answer whether a similar scheme can be realized in the robot learning field, i.e.,
\begin{align} \label{eq_objective}
	\bm{\hat f}(t) = \bm{f}_{\bm{\theta}}(t) + \mathcal{F}(\bm{\gamma}(t))
\end{align}
where $\bm{\hat f}(t)$ represents the corrected learning outcome and $\mathcal{F}(\cdot)$ denotes a user-defined low-pass filter. If Eq. \eqref{eq_objective} can be achieved, the low-frequency characteristic of learning residual can be captured convergently and compensated timely, enabling the refinement of learning outcomes. In this sense, the convergence process of the low-pass filter can be regarded as an adaptive procedure in unseen scenarios, like living organisms in nature. In this work, we show that the objective in Eq. \eqref{eq_objective} can be fully realized by utilizing real-time state feedback and \textit{a priori} structure of the differential equation, such as kinematics or dynamics. Its implementation is detailed in the ``Filtering learning residual'' section in Methods. We also show that the filter parameters can be auto-learned from an optimal-control perspective (``Learning filter parameter'' section in Methods).


\subsection{{Learning policy for quadruped locomotion}}

\begin{figure*}
	\centering
	\includegraphics[width=1\linewidth]{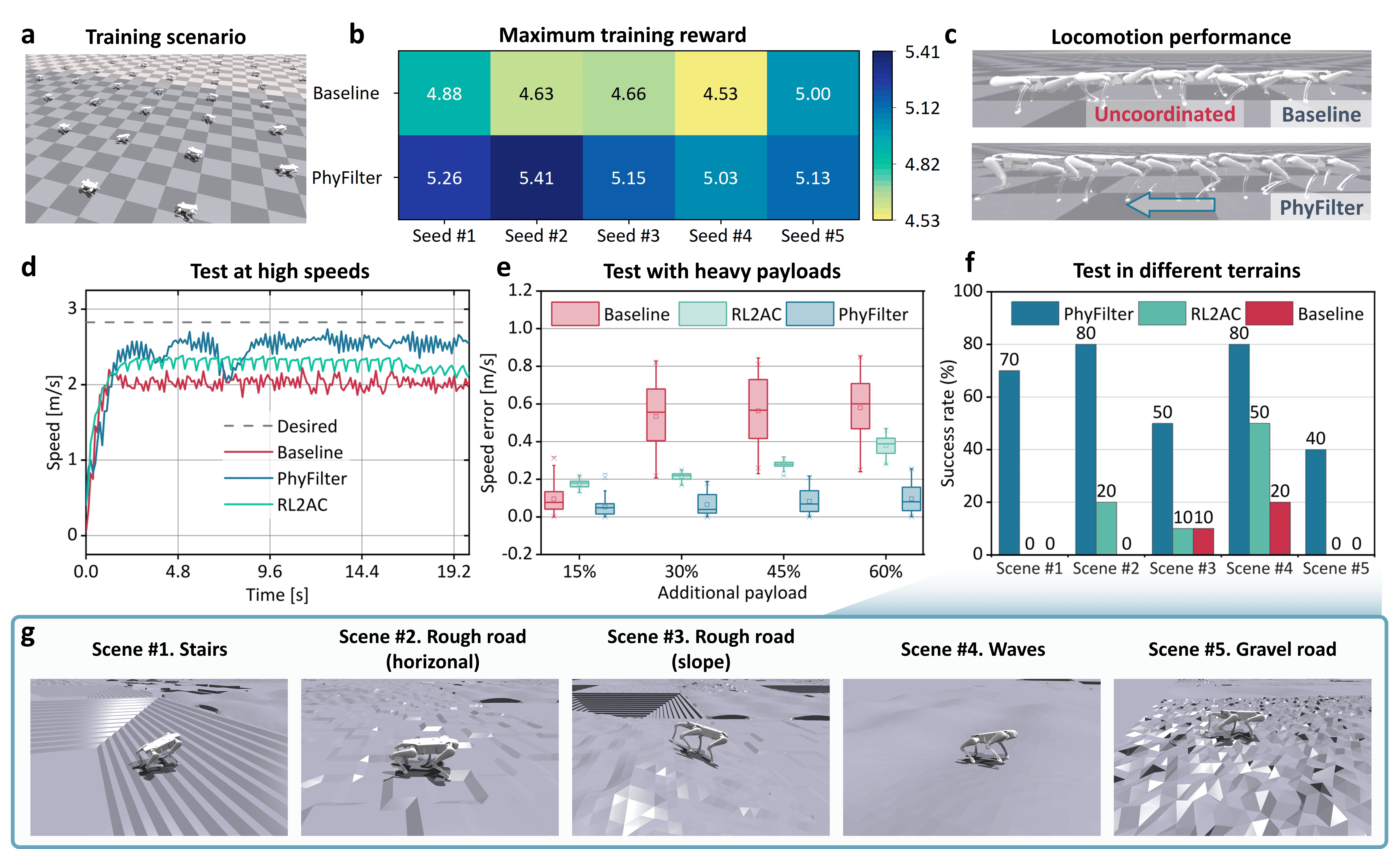}
	\caption{\textbf{{Simulation results of the quadruped example.}} \textbf{a}, Parallel training setup using only flat terrain. \textbf{b}, Maximum training rewards over five random seeds. The number of iteration and training environment are set as $15000$ and $256$, respectively. \textbf{c}, Locomotion performance of the baseline and PhyFilter in the training environment. \textbf{d}, Testing performance at an unseen high speed of $2.8\ \text{m/s}$. \textbf{e}, Testing performance under unseen payloads ranging from $15\% \sim 60\%$ of the robot’s $11.86\ \text{kg}$ weight. \textbf{f}, Testing performance across five unseen scenarios. \textbf{g}, Stairs, rough, wave, and gravel roads, unseen in training, are considered during deployment.}
	\label{quadruped__simulation_result}
\end{figure*}

\begin{figure*}
	\centering
	\includegraphics[width=1\linewidth]{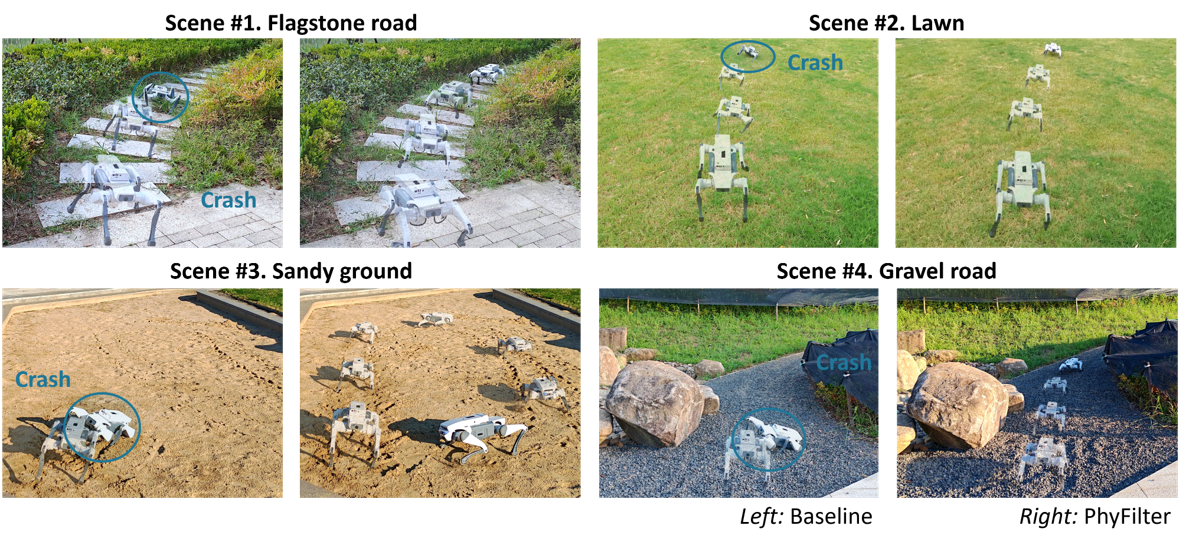}
	\caption{\textbf{Experimental results of the quadruped example.} Several unseen scenarios including flagstone, lawn, sand, and gravel terrains are evaluated. Notably, the policy is trained solely on simulated flat terrain with no post-deployment tuning. }
	\label{quadruped__experiment_result}
\end{figure*}

Locomotion control of quadruped robots has garnered substantial interest in recent years. Notably, RL-based methods have achieved impressive robustness on challenging terrains by virtue of parallel training and domain randomization strategies~\citep{han2024lifelike, yang2020multi, kumar2021rma, lee2024learning, kim2025high, shi2024rethinking, rudin2022learning}. In this paper, we will show that incorporating PhyFilter into the RL training pipeline not only increases maximum training rewards but also significantly enhances generalization to diverse, previously unseen real-world environments, even when trained only in parsimonious scenarios.

The parallel deep learning training strategy presented in~\citep{rudin2022learning} is adopted as the baseline. Desired joint positions are inferred from proprioceptive observations and body commands, while the proposed PhyFilter is integrated into the joint-space control layer. Additional implementation details are provided in the ``Implementation details of quadruped example'' section in Methods. \textcolor{black}{To emphasize the generalization ability of the PhyFilter-enhanced RL scheme, the training environment in simulation only considers a flat terrain (Fig. \ref{quadruped__simulation_result}a) and standard parameter randomization ranges (e.g., $-1\sim1\ \text{m/s}$ velocity at each DoF, $-1\sim3\ \text{kg}$ payload)~\citep{deeprobotics_rltraining,long2023hybrid}. Additional tests trained under full randomized terrains are provided in Supplementary Section I and Fig. S5.} During employment, however, the robot encounters broader disturbances, including unseen internal dynamical parameters and diverse unstructured terrains. This setting highlights how the proposed method reduces the randomization burden of conventional RL while maintaining remarkable generalization ability.

Fig. \ref{quadruped__simulation_result}b shows the training performance from five random seeds for both the baseline and the proposed method. PhyFilter consistently achieves higher maximum rewards. Fig. \ref{quadruped__simulation_result}c further illustrates that PhyFilter exhibits a more coordinated locomotion, whereas baseline displays degraded gait quality. This improvement likely arises from the filtering mechanism, which effectively compensates for unknown uncertainties and facilitates policy training. This observation suggests that incorporating a robust low-level controller can substantially enhance the learning efficiency of high-level RL policies.

Subsequently, we further evaluate generalization during deployment under high speed ($2.83\ \text{m/s}$), varying payloads ($15\% \sim 60\%$ of robot weight $11.86\ \text{kg}$, $137.33\%$ beyond the training-set maximum), and diverse unseen terrains (stairs, rough, wave, and gravel roads, Fig. \ref{quadruped__simulation_result}g). Besides the baseline, RL2AC~\citep{lyu2024rl2ac}, which combines an adaptive controller with RL to mitigate the sim-to-real gap, is also compared. For speed tracking (Fig. \ref{quadruped__simulation_result}d), PhyFilter achieves superior accuracy relative to the baseline and RL2AC. With respect to payload tests (Fig. \ref{quadruped__simulation_result}e), the tracking performance of the baseline and RL2AC degrades with increasing payload, whereas PhyFilter maintains a smaller tracking error ($\le 0.2\ \text{m/s}$) even under $60\%$ of robot weight payload. For terrain adaptation (Fig. \ref{quadruped__simulation_result}f), since only flat terrain is considered during training, the baseline struggles to adapt to new terrains as expected. However, the proposed algorithm exhibits excellent terrain adaptation. Across five different scenarios, the passing success rates of the proposed algorithm are consistently higher than those of the baseline and RL2AC, even though these scenarios have never been seen before.

Finally, the trained policy is directly deployed on the real hardware (Fig. S4) to evaluate the sim-to-real transfer performance, with a particular focus on unseen terrains. As shown in Fig. \ref{quadruped__experiment_result}, the quadruped traverses four challenging natural terrains without additional tuning. The baseline exhibits poor adaptability and collapses immediately on sandy and gravel surfaces. In stark contrast, the PhyFilter-enhanced policy enables stable locomotion across all tested terrains, despite being trained exclusively on simulated flat ground. \textcolor{black}{Moreover, the PhyFilter-guided policy can also transfer successfully across real terrains even with the correction switched off at deployment, which reveals PhyFilter shapes an intrinsically better policy rather than merely compensating at runtime.} To recap, PhyFilter can not only improve the RL training performance, but also dramatically enhance its generalization ability using simplified physical knowledge (``Implementation details of quadruped example'' section in Methods).

\subsection{Learning dynamics for drone maneuvering flight}

\begin{figure*}
	\centering
	\includegraphics[width=0.8\linewidth]{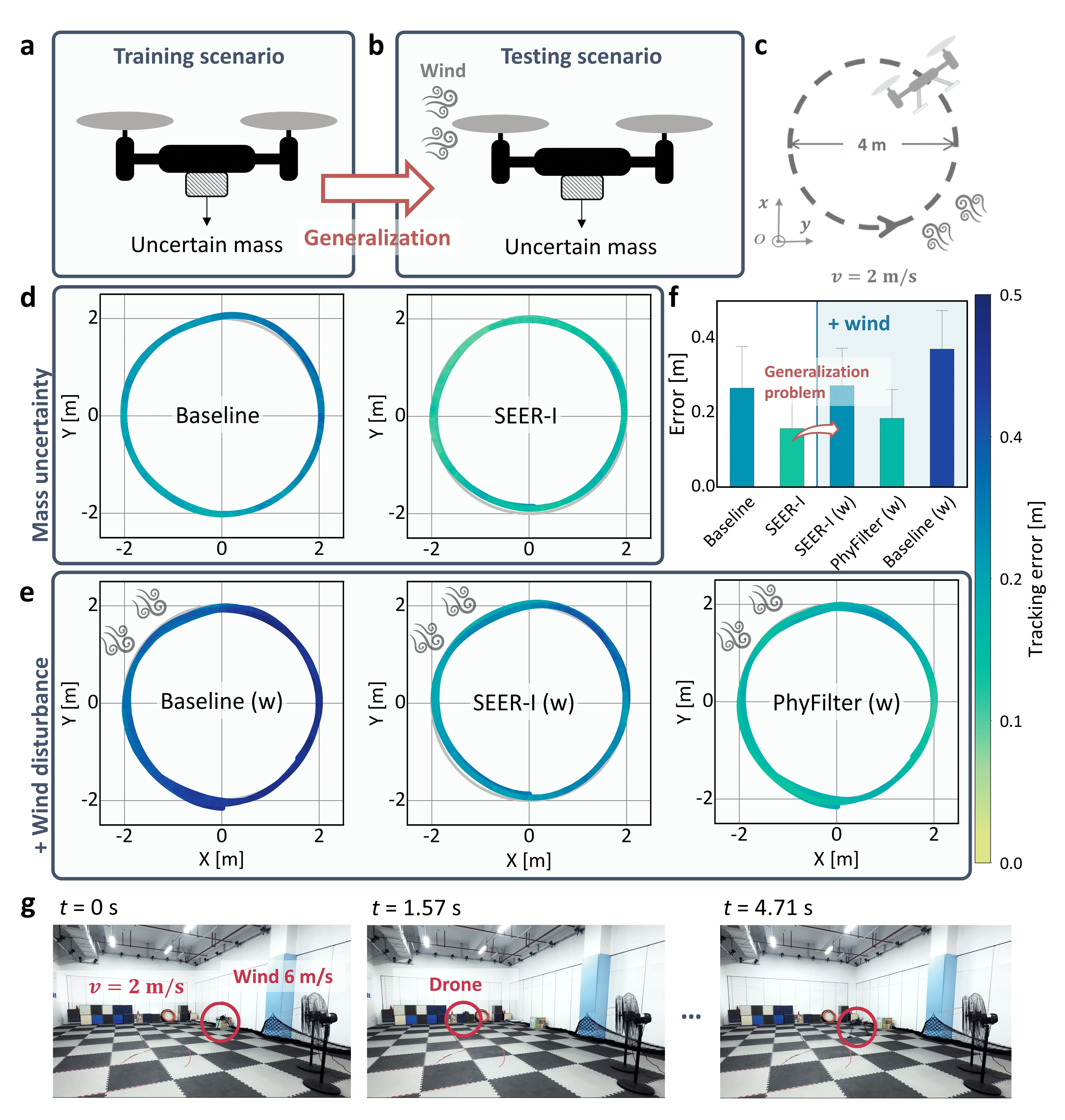}
	\caption{\textbf{Experimental results of the drone maneuvering example.} \textbf{a}, The employed SEER-I is trained under mass uncertainty, while the testing scenario further includes the wind disturbance, as depicted in \textbf{b}. \textbf{c}, In tests, the drone is commanded to follow a circular trajectory with $2\ \text{m/s}$. \textbf{d}, The tracking performance with only mass uncertainty under the baseline and SEER-I. \textbf{e}, The tracking performance with both mass uncertainty and wind disturbance under the baseline, SEER-I, and PhyFilter. \textbf{f}, Mean absolute tracking errors across all implemented tests. The error bar indicates the tracking errors presented in (\textbf{d}) and (\textbf{e}). \textbf{g}, Several flight snapshots.}
	\label{dynamics_result_circle_error}
\end{figure*}

Beyond RL, SL has also been widely applied across diverse robotic domains. In our second case, we integrate PhyFilter into an SL-based controller on a drone to evaluate its ability to enhance generalization under previously unseen disturbances. Notably, PhyFilter is used in a plug-and-play manner in this case. Mass uncertainties usually appear in payload-transport missions for drones. SEER-I, a learning-based adaptive estimator developed by~\citep{jia2025}, can effectively estimate mass uncertainty by combining offline basis learning with online adaptive control. However, the SEER-I generalizes poorly to other kinds of unseen disturbances, such as wind. In this part, we augment SEER-I with physics knowledge to assess how PhyFilter handles unseen disturbances. More details about the implementation of SEER-I on the drone can be found in the ``Implementation details of drone flight example'' section in Methods.

In tests, the drone is commanded to follow a circular trajectory with $2\ \text{m/s}$ (Fig. \ref{dynamics_result_circle_error}c) under different uncertainties. We first assess the learned SEER-I model under a single mass uncertainty. As shown in Fig. \ref{dynamics_result_circle_error}d, the SEER-I can achieve better tracking performance compared with the baseline~\citep{9779449}. However, when the wind disturbance is further included, from Fig. \ref{dynamics_result_circle_error} (e and f), the tracking performance of SEER-I is degraded, which indicates its limited ability to generalize to unseen uncertainties. {After the presented PhyFilter is employed, the mean absolute tracking error can be reduced by $30.22\%$ relative to SEER-I and $50.17\%$ relative to the baseline.} Representative flight snapshots are shown in Fig. \ref{dynamics_result_circle_error}g. The uncertainty estimation performance of SEER-I and PhyFilter is further provided in Supplementary Fig. S1, which shows that SEER-I captures only mass variations, whereas PhyFilter additionally identifies wind-induced and inherent dynamical uncertainties. To sum up, these results suggest that PhyFilter can improve the generalization ability of SL models by leveraging real-time feedback state and dynamical structure.

\subsection{{Learning dynamics for aerial manipulation}}

\begin{figure*}
	\centering
	\includegraphics[width=1\linewidth]{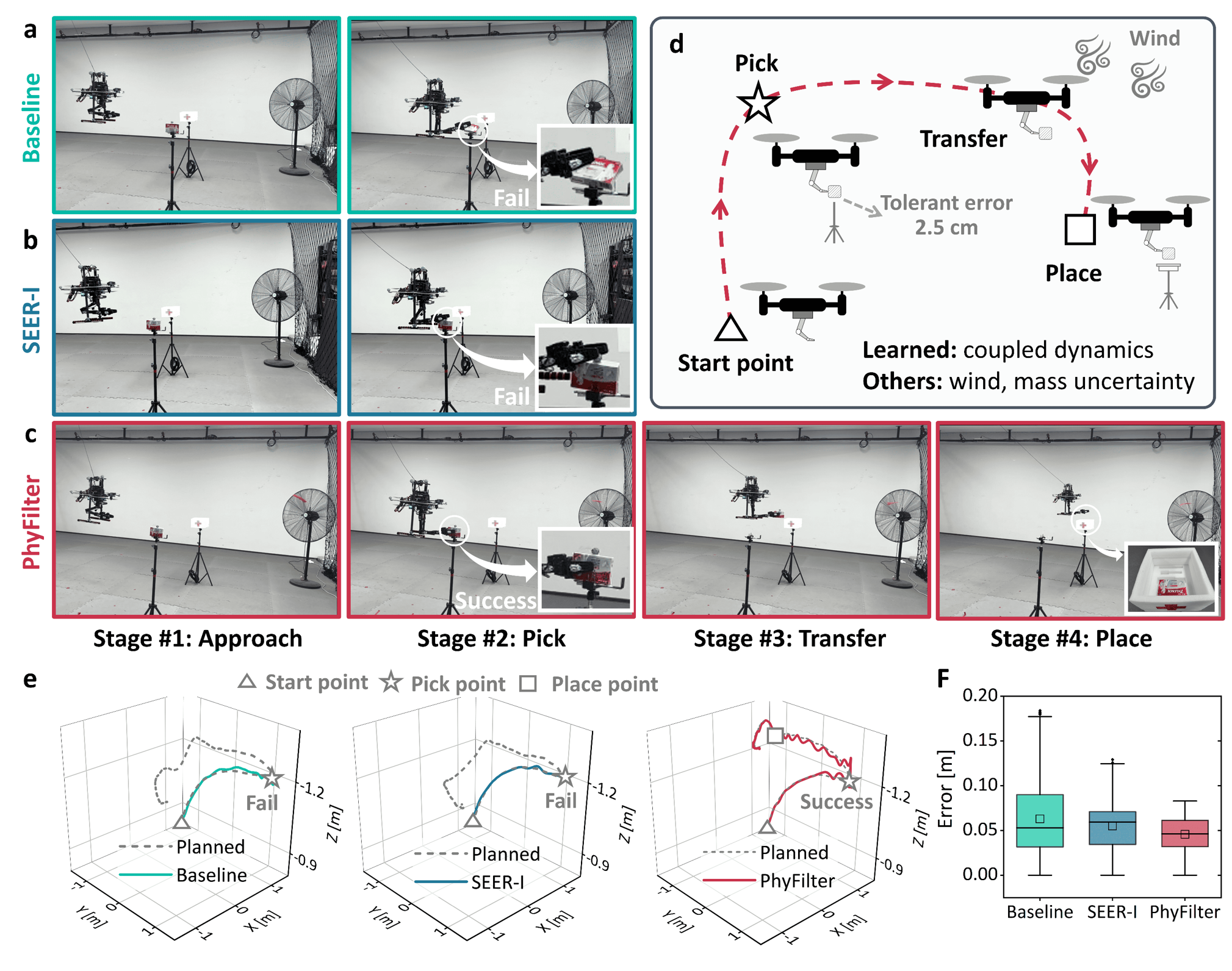}
	\caption{\textbf{Experimental results of the aerial manipulation example.} \textbf{a}-\textbf{c}, Pick-and-place executions under the baseline, SEER-I, and PhyFilter, respectively. \textbf{d}, Illustration of the pick-and-place task involving coupled dynamics, wind disturbance, and mass uncertainty. \textbf{e}, Three-dimensional tracking performance of the compared methods. \textbf{f}, Tracking errors for each method.}
	\label{manipulation__experiment_result}
\end{figure*}

Compared with conventional multirotor platforms, aerial manipulators offer substantially greater flight flexibility and operational dexterity, enabling deployment in complex interactive tasks such as component transport, infrastructure repair, and obstacle removal~\citep{9462539}. High-precision end-effector control is a fundamental prerequisite for the deployment of aerial manipulators. However, achieving such precision is challenging due to the presence of strong coupling dynamics caused by the manipulation movement, along with external disturbances ~\citep{9476743,9462539, SR_PHRI}. 

To evaluate the compatibility of PhyFilter in such precision-critical interaction scenarios, we design a pick-and-place mission. As illustrated in Fig. \ref{manipulation__experiment_result}d, the aerial manipulator is expected to precisely pick a medicine from a tripod and place it into the specified area. The maximum allowable tracking error of the end-effector is $2.5$ cm. The pick-and-place reference trajectory is computed in real time using a model predictive control strategy (``Trajectory generation of aerial manipulation'' section in Supplementary Information). To mimic realistic environmental disturbances, a $380$~W fan is used to generate a maximum wind speed of up to $5$ m/s. Additionally, an unmodeled $10\%$ mass uncertainty ($0.3$ kg) is introduced to replicate real-world payload change. These factors constitute a fundamental active interaction task in the emergency rescue field.

Fig. \ref{manipulation__experiment_result} (a-c) presents representative snapshots corresponding to three distinct control schemes: the baseline controller, SEER-I~\citep{jia2025}, and the proposed PhyFilter. SEER-I is an offline-trained learning-based model that captures coupled aerial manipulation dynamics using collected states and ground-truth disturbance labels. Implementation details for the baseline and SEER-I are provided in the ``Implementation details of aerial manipulation example'' section in Methods. PhyFilter is applied on top of SEER-I to enhance generalization to unseen disturbances, including external wind and mass uncertainty. The resulting three-dimensional (3D) tracking performance and tracking errors of these methods are visualized in Fig. \ref{manipulation__experiment_result} (e and f). As observed from the experimental results, both the baseline and SEER-I fail to capture the object due to the involvement of unseen wind and mass uncertainties. Notably, owing to its incorporation of the coupled dynamics model, SEER-I exhibits marginally superior tracking performance compared to the baseline. In stark contrast, PhyFilter markedly improves the generalization capability of SEER-I, enabling the aerial manipulator to maintain centimeter-level manipulation and perform the pick-and-place task.

\subsection{Learning kinematics for acceleration perception}

\begin{figure*}
	\centering
	\includegraphics[width=0.8\linewidth]{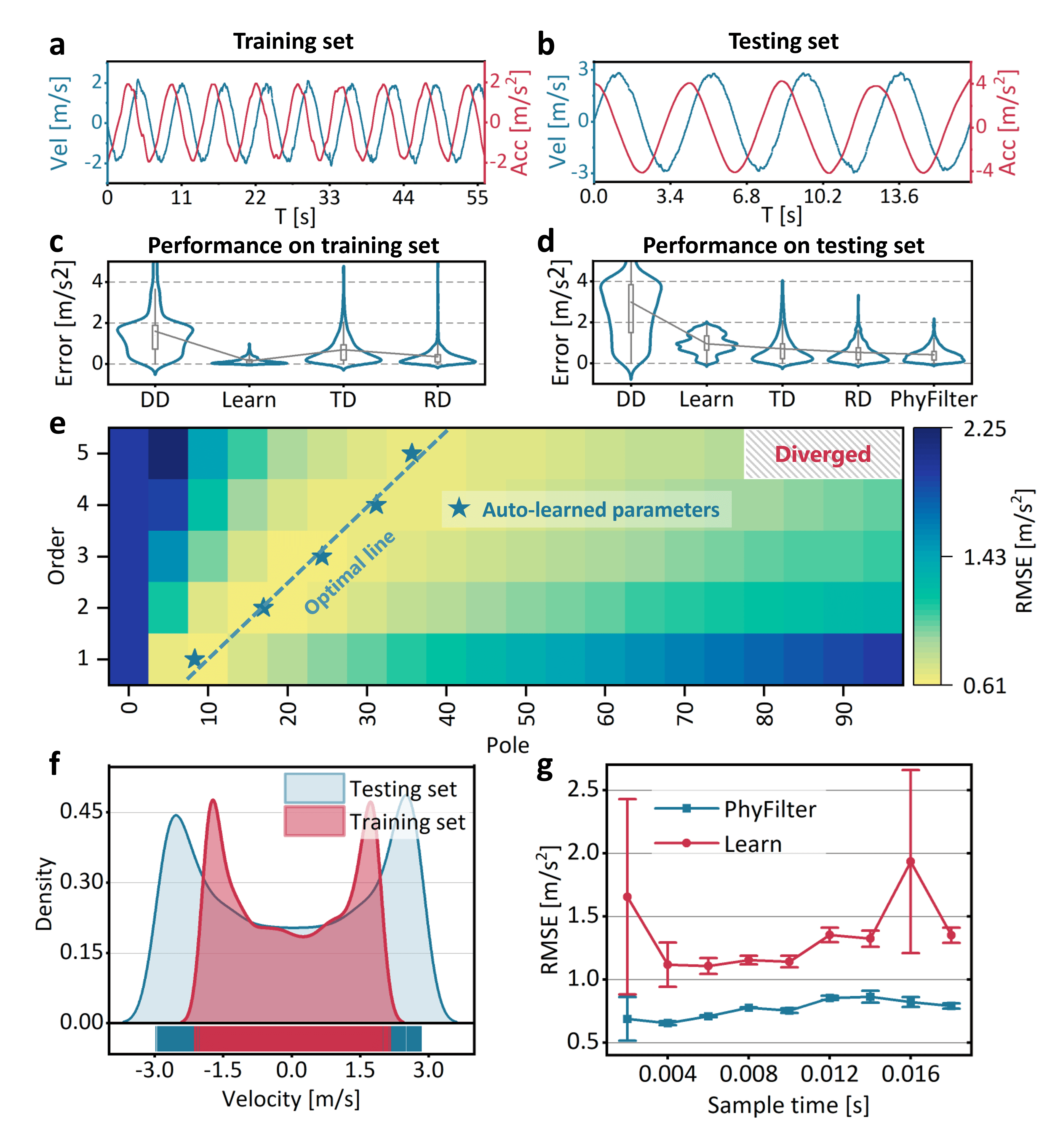}
	\caption{\textbf{Experimental results of the accelerator example.} \textbf{a}, Training dataset containing velocities from $0$ to $2\ \text{m/s}$. \textbf{b}, Testing dataset containing velocities from $0$ to $3\ \text{m/s}$. \textbf{c}, Absolute prediction errors of each method on the training set. \textbf{d}, Absolute prediction errors on the testing set. \textbf{e}, Ablation study on filtering orders and parameters determined via \textit{pole-placement}. Blue stars denote the auto-learned parameters, which are roughly consistent with manually optimal results. \textbf{f}, Distribution of the training and testing datasets. \textbf{g}, Ablation study across different sampling intervals. RMSE denotes the root mean square error.}
	\label{perception_result0}
\end{figure*}

Precisely estimating robot acceleration is particularly challenging for resource-constrained platforms such as lightweight drones, which typically lack additional high-fidelity perceptive sensors~\citep{jia2025}. The neural network-based differentiators have shown promising performance for state perception~\citep{9635983}. However, as with other neural network-based application scenarios, the generalization problem is an intractable difficulty that cannot be bypassed. In this section, we apply PhyFilter to enhance the robustness and generalization of learned acceleration estimates. Additional theoretical and implementation details are provided in the ``Implementation details of perception example'' section in Methods.

The training set ($0\sim2\ \text{m/s}$) and testing set ($0\sim3\ \text{m/s} $) are collected from real flight experiments, as shown in Fig. \ref{perception_result0} (a, b, and f). A fully connected neural network with two hidden layers is first trained in a supervised manner to map historical velocities to acceleration. Subsequently, PhyFilter using the knowledge of kinematic structure, is employed to enhance the learning output in the face of the testing set with distribution shift. Several state-of-the-art benchmarks, including robust differentiator (RD)~\citep{seeber2021robust} with a predefined convergence time, tracking differentiator (TD)~\citep{han2009pid}, and discrete differentiator (DD) are also implemented as compared methods. All baselines are carefully tuned for their best performance~\citep{10288520}. As shown in Fig. \ref{perception_result0} (c and d), the purely learning-based model exhibits generalization failure in the testing set. However, it can be seen that PhyFilter achieves the highest accuracy, outperforming both learning-based and classical differentiator baselines.

Ablation experiments on filter order and pole selection (“Filtering learning residual” section in Methods) are summarized in Fig. \ref{perception_result0}e. The results indicate that higher-order filters benefit from larger pole values. Moreover, an auto-learning procedure (“Learning filter parameter” section in Methods) is further developed to discover the optimal filter parameters. As can be observed from Fig. \ref{perception_result0}e, the auto-learned optimal parameters are roughly consistent with the manually adjusted ones, showing the effectiveness of the proposed strategy. Since differentiator performance depends on sampling rate, we additionally examine a range of sampling intervals, as presented in Fig. \ref{perception_result0}g. The proposed PhyFilter consistently outperforms the purely learning-based one, ranging from $0.002\ \text{s}$ to $0.018\ \text{s}$ in the testing set. 

\section{Discussion} \label{sec_diss}

Stemming from the adaptive mechanisms observed in living organisms, we presented a physics-filtered learning framework aiming at alleviating the long-standing generalization challenge in robot learning. To achieve low-pass filtering of learning residuals (Eq. \eqref{eq_objective}), mimicking biological systems, we derived an equivalent implementation (Eq. \eqref{eq_observer}) that leverages real-time state feedback and the intrinsic differential structure of physical models. To eliminate the need for manual parameter tuning, an automated learning algorithm was further developed. Overall, PhyFilter exhibited the advantages of generalization and interpretability.

Four realistic robotic experiments were arranged to validate the effectiveness of PhyFilter. When incorporated with RL strategies for quadrupedal locomotion, PhyFilter not only enhanced training rewards but also substantially improved generalization. Notably, even if the RL policy was trained solely on simulated flat terrain, it could stably generalize to unseen realistic terrains. Additionally, PhyFilter boosted the generalization capability of SL methods. Flight tests demonstrated that, with PhyFilter, drones and aerial manipulators maintained satisfactory control precision while encountering unseen uncertainties. We also demonstrated that PhyFilter also strengthened perception modules, enabling accurate acceleration estimation despite input distribution shifts. Note that PhyFilter is applied in a plug-and-play manner in the last three experiments, underscoring its model-agnostic feature.

\textcolor{black}{To bridge the simulation-to-reality gap, the prevailing paradigm relies on domain randomization~\citep{tobin2017domain, tobin2018domain}, which stochastically randomizes the parameters of the simulator to encompass real-world uncertainties. This approach pursues the average performance among randomized environments, i.e., sacrificing accuracy for robustness~\citep{o2022neural, jia2024feedback, ha2024learning, green2012linear}, similar to classical robust control~\citep{ha2024learning, green2012linear}. In contrast, without excessive domain randomization, PhyFilter can unify generalization and accuracy in one framework via a physics-informed feedback mechanism~\citep{jia2024feedback}. Specifically, when encountering scenarios outside the training distribution, PhyFilter is activated to enhance generalization. Conversely, within familiar training regimes, where learning residuals remain small, its intervention is adaptively attenuated, thereby preserving the intrinsic precision in normal cases.}

\textcolor{black}{The enhanced learning strategy no longer requires extensive domain randomization in the training phase while still maintaining excellent generalization during deployment. In the quadruped experiment, PhyFilter eliminates the need to account for complex terrains, payloads, and speeds during training, avoiding substantial engineering effort and repeated manual tuning for constructing effective randomization parameters.} During deployment, PhyFilter required only a small set of analytic update equations (Fig. \ref{All_scenarios}b). In the cases of drone flight and aerial manipulation, PhyFilter ran in real time on an STM32F765 microprocessor at $500$ Hz, demonstrating its suitability for lightweight, resource-constrained platforms.

Three limitations exist in this study that also highlight directions for future development. First, the beneficial knowledge obtained through PhyFilter only enhances the learning output~\citep{lyu2026adapt}. A more intelligent strategy should involve leveraging the filtered knowledge to refine the parameter or structure of the original neural network. Continual learning~\citep{parisi2019continual} may be an integrable approach to achieve this goal. \textcolor{black}{Secondly, while the parameter auto-learning algorithm converges reliably without manual tuning, each iteration requires a full forward rollout of the trajectory, which limits its efficiency on large-scale data. Improving its computational cost, for instance through truncated or parallelized rollouts, is left as future work.} \textcolor{black}{Lastly, apart from the neural-network prediction error, $\bm{\gamma}(t)$ may also lump other uncertainties, such as the modeling mismatch in the quadruped case. Separating these components, possibly through meta-learning~\citep{jia2025}, is left as future work.}




\section{Methods} \label{sec_method}

\subsection{Filtering learning residual} \label{sec_theorem}

In this part, the theoretical framework of  PhyFilter is established. Before proceeding, consider a general differential structure
\begin{align} \label{eq_model}
	\bm{\dot{x}} = \bm{f}(\bm{x}, \bm{z}) + \bm{g}(\bm{x}, \bm{z})
\end{align} 
with internal system state $\bm{x} \in \mathbb{R}^{n_x}$, \textit{optional} external influencing factor $\bm{z} \in \mathbb{R}^{n_z}$, known nonlinear mapping $\bm{g}(\cdot): \mathbb{R}^{n_x} \times \mathbb{R}^{n_z} \rightarrow \mathbb{R}^{n_x}$, and unknown nonlinear part $\bm{f}(\cdot): \mathbb{R}^{n_x} \times \mathbb{R}^{n_z} \rightarrow \mathbb{R}^{n_x}$. Note that the above structure can represent unknown dynamics and kinematics in robotic systems. \textcolor{black}{Unless otherwise specified, $\bm{g}(\bm{x}, \bm{z})$ and $\bm{f}(\bm{x}, \bm{z})$ are abbreviated as $\bm{g}(t)$ and $\bm{f}(t)$, respectively,} by considering the time-varying features of $\bm{x}(t)$ and $\bm{z}(t)$. It is assumed that $\bm{f}(t)$ can be learned through neural networks or kernel-based regression methods with parameter $\bm{\theta} \in \mathbb{R}^{n_\theta}$, denoted as $\bm{f}_{\bm{\theta}}(t)$, with learning residual $\bm{\gamma}(t) \in \mathbb{R}^{n_x}$, i.e., $\bm{\gamma}(t) = \bm{f}(t) - \bm{f}_{\bm{\theta}}(t)$. \textcolor{black}{Note that besides the neural network prediction error, $\bm{\gamma}(t)$ can also lump analytical modeling error of $\bm{g}(t)$ and other uncertainties, all of which are treated as the quantity to be suppressed.}

Next, we show that the objective in Eq. \eqref{eq_objective} can be realized from the filtering perspective. The direct implementation of Eq. \eqref{eq_objective} is not feasible because $\bm{\gamma}(t)$ cannot be accessed without additional sensing. However, by leveraging available physical structure and real-time state feedback, we demonstrate that Eq. \eqref{eq_objective} can be sufficiently achieved. The proposed formulation can also be extended to more general classes of filters. More details can be found in the ``Filtering learning residual with more general form'' section in Supplementary Information. 

Consider an $n$th-order low-pass filter with the following transfer function
\begin{align} \label{eq_n_filter}
	\mathcal{L} \left[\mathcal{F}(\bm{\gamma}(t))\right] = \frac{a_0}{s^n + a_{n-1}s^{n-1} + \cdots + a_{1}s + a_0}{\Upsilon}(s)
\end{align}
with \textit{Laplace} transform $\mathcal{L}\left[\cdot\right]$, \textit{Laplace} variable $s$, parameters $a_{n-1}, \cdots, a_{1}, a_{0}$, and frequency-domain expression ${\Upsilon}(s)$ of $\bm{\gamma}(t)$. By substituting Eq. \eqref{eq_n_filter} into Eq. \eqref{eq_objective} and performing the result in time-domain, it can be rendered that 
\begin{align*}
	& \bm{\hat{f}}^{(n)}(t) + a_{n-1}\bm{\hat{f}}^{(n-1)}(t) + \cdots + a_1 \bm{\hat{f}}{'}(t) + a_0 \bm{\hat{f}}(t) \notag\\ = & \bm{f}_{\bm{\theta}}^{(n)}(t) + a_{n-1}\bm{f}_{\bm{\theta}}^{(n-1)}(t) + \cdots + a_1 \bm{f}_{\bm{\theta}}'(t) + a_0 \bm{f}_{\bm{\theta}}(t) + a_0 \bm{\gamma}(t)
\end{align*}
where $(\cdot)^{(n)}$ denotes the $n$th-order differentiation of a signal. For simplicity,
define $\bm{\chi}_1(t) = {a_{n - 1}}\int {\bm {\hat{f}}(t)dt}  +  \cdots + {a_1}{\int }^{n - 1}\bm{\hat {f}}(t)(dt)^{n-1}$ and $\bm{\chi}_2(t) = \bm{f}_{\bm{\theta}}(t) + {a_{n - 1}}\int {\bm {{f}}_{\bm{\theta}}(t)dt}  +  \cdots + {a_1}{\int}^{n - 1}\bm{{f}}_{\bm{\theta}}(t)(dt)^{n-1}$, where ${\int}^{i}(\cdot)(dt)^i$ denotes the $i$th-order integration of a signal. Thus, we have
\begin{align*}
	\bm{\hat f}^{(n)}(t) + \bm{\chi}^{(n)}_1 + a_0 \bm{\hat{f}}(t) = \bm{\chi}_2^{(n)} + a_0 \bm{f}_{\bm{\theta}}(t) + a_0 \bm{\gamma}(t).
\end{align*}
Due to that $\bm{\gamma}(t) = \bm{f}(t) - \bm{f}_{\bm{\theta}}(t)$ and physical structure in Eq. \eqref{eq_model}, one can imply
\begin{align} \label{eq_mid1}
	\bm{\hat f}^{(n)}(t) + \bm{\chi}_1^{(n)} + a_0 \bm{\hat{f}}(t) = \bm{\chi}_2^{(n)} + a_0 \bm{f}_{\bm{\theta}}(t) + a_0 (\bm{\dot{x}}-  \bm{g}(t) - \bm{f}_{\bm{\theta}}(t)).
\end{align}
It can be seen $\bm{\hat f}^{(n)}(t)$, $\bm{\chi}_1^{(n)}$, $\bm{\chi}_2^{(n)}$, and $\bm{\dot{x}}$ in Eq. \eqref{eq_mid1} are unknown or strenuous to be accurately obtained. Define an auxiliary variable $\bm{\xi} = (\bm{\hat f}(t) + \bm{\chi}_1- \bm{\chi}_2 - a_0 {\int}^{n-1}\bm{x}(dt)^{n-1})/a_0$, leading to $a_0\bm{\xi}^{(n)} = \bm{\hat f}^{(n)}(t) + \bm{\chi}_1^{(n)} - \bm{\chi}_2^{(n)} - a_0\bm{\dot{x}}$. Thus, it can be obtained that
\begin{align} \label{eq_observer}
	\left\{ {\begin{array}{*{20}{c}}
			{\bm{\xi}^{(n)} = -  \bm{g}(t) - \bm{\hat{f}}(t)}\\
			{\bm{\hat{f}}(t) = a_0\bm{\xi} - \bm{\chi}_1 + \bm{\chi}_2 +  a_0 {\int}^{n-1}\bm{x}(dt)^{n-1}.}
	\end{array}} \right.
\end{align}

By introducing a new variable $\bm{\xi}$, the direct use of $\bm{\hat f}^{(n)}(t)$, $\bm{\chi}_1^{(n)}$, $\bm{\chi}_2^{(n)}$, and $\bm{\dot{x}}$ in Eq.~\eqref{eq_mid1} is avoided. The underlying principle is to reconstruct lower-order signal evolution by combining higher-order signal differences with lower-order initial conditions. The resulting analytic implementation of Eq. \eqref{eq_observer} is portrayed in Fig. \ref{All_scenarios}b. 

Up to now, the objective stated in Eq. \eqref{eq_objective} has been reformulated into the implementable form of Eq. \eqref{eq_observer}, relying solely on accessible measurements. In the case where the low-pass filter is configured as a first-order one, Eq. \eqref{eq_observer} will roughly equal to EVOLVER~\citep{10288520}, SEER-II~\citep{jia2025}, and feedback neural network~\citep{jia2024feedback, an2026feedback}, in which the \textit{Koopman} operator, \textit{Chebyshev} polynomials, and neural ordinary differential equations (neural ODEs)~\citep{chen2018neural} are respectively utilized to learn $\bm{f}(t)$ with specific profitable characteristics (``Several special forms'' section in Supplementary Information). \textcolor{black}{Furthermore, if the learned knowledge is additionally disregarded under a first-order setting, Eq. \eqref{eq_observer} ultimately degenerates to a purely feedback-based disturbance observer~\citep{NDO_constant, DOBC_survey}.}

\textcolor{black}{The feedback mechanism of PhyFilter is originally inspired by the disturbance observer~\citep{NDO_constant, DOBC_survey}, upon which it makes several advances: during RL training, PhyFilter guides the generation of a better policy rather than acting only at deployment; it is integrated with the learning output in SL-based cases, reducing the conservatism of a classical disturbance observer that treats all analytically unmodeled dynamics as a single lumped disturbance; it admits higher-order forms in the filtering aspect to handle more complex residuals beyond the first-order case; and its gains can be auto-learned instead of manually tuned. It is this combination, not the disturbance-observer backbone alone, that constitutes PhyFilter.}

\subsection{\textcolor{black}{Learning filter parameter}}
\label{sec_learning_main}

\textcolor{black}{The filter parameters $\bm{a}_f = \{a_0, a_1, \cdots, a_{n-1}\}$ can be set by the mature \textit{pole-placement} theory, but manually placing the poles becomes tedious and non-intuitive as the filter order grows. To this end, we further develop an auto-learning algorithm that determines $\bm{a}_f$ from an optimal-control perspective. Herein we provide only an outline. The complete derivation is given in the “Formalization of parameter learning” section in Supplementary Information.}

\textcolor{black}{The key idea is to reinterpret the filter parameters $\bm{a}_f$ as the \emph{control input} of a dynamical system, whose state $\bm{\varsigma}$ concatenates the auxiliary variable $\bm{\xi}$ together with the successive higher-order integrals of $\bm{\hat{f}}$ and $\bm{f}_{\bm{\theta}}$ (defined in the Supplementary Information). Rearranging Eq.~(5) casts the filter as the discrete dynamics $\bm{\varsigma}_{i+1} = \bm{\phi}^d_{\bm{\varsigma}}(\bm{\varsigma}_i, \bm{a}_f)$, so that tuning $\bm{a}_f$ becomes the optimal-control problem}
\begin{align} \label{eq_optimal_obj_main}
	\textcolor{black}{\mathop{\min}\limits_{\bm{a}_f} J(\bm{a}_f)
		= \sum_{i=1}^{N-1} l_i(\bm{f}_i^*, \bm{\hat{f}}_i, \bm{a}_f) + l_N(\bm{f}_N^*, \bm{\hat{f}}_N),
		\quad \text{s.t.}\ \ \bm{\varsigma}_{i+1} = \bm{\phi}^d_{\bm{\varsigma}}(\bm{\varsigma}_i, \bm{a}_f),}
\end{align}
\textcolor{black}{where $N$ is the sample number and the stage cost $l_i(\cdot) = (\bm{f}_i^* - \bm{\hat{f}}_i)^\top(\bm{f}_i^* - \bm{\hat{f}}_i)$ penalizes the discrepancy between the model rollout $\bm{\hat{f}}_i$ and the labelled sample $\bm{f}_i^*$.}

\textcolor{black}{Applying the \textit{Lagrange} multiplier method to Eq.~\eqref{eq_optimal_obj_main} yields the first-order optimality conditions, which take the form of a costate recursion together with a stationarity condition on $\bm{a}_f$. Solving these conditions directly is expensive for large $N$ and state dimension, so we instead compute the analytic gradient $\nabla_{\bm{a}_f}\mathcal{L}$ by a forward rollout of the state followed by a backward \emph{adjoint solve} (reverse-mode differentiation)~\citep{chen2018neural}, and update $\bm{a}_f$ by gradient descent. Mini-batching, the \textit{Adam} optimizer, and early stopping are used to improve efficiency and stability. The full optimality conditions and the two resulting algorithms are detailed in Supplementary Information. The learning performance is verified on the acceleration example (Fig. \ref{perception_result0}e), with additional results in Supplementary Fig.~S2.}      

\subsection{Implementation details of quadruped example} \label{method_quadruped}

The fully-actuated \textit{Lagrangian} dynamical model of the quadruped robot is usually described as 
\begin{align} \label{eq_quadruped_model}
	\bm{M}(\bm{q})\ddot{\bm{q}} + \bm{C}(\bm{q}, \bm{\dot{q}})\bm{\dot{q}} + \bm{G}(\bm{q}) -\bm{J}^T\bm{F}_c = \bm{\tau} + \Delta \bm{\tau}
\end{align}
with joint angle $\bm{q} \in \mathbb{R}^{12}$, inertial matrix $\bm{M}(\bm{q}) \in \mathbb{R}^{{12} \times {12}}$, \textit{Coriolis} force $\bm{C}(\bm{q}, \bm{\dot{q}})\in \mathbb{R}^{{12} \times {12}}$, gravity force $\bm{G}(\bm{q}) \in \mathbb{R}^{12}$, \textit{Jacobian} matrix $\bm{J} \in \mathbb{R}^{{12} \times 3}$, contact force $\bm{F}_c \in \mathbb{R}^{3}$, joint torque input $\bm{\tau}  \in \mathbb{R}^{12}$, and unknown torque uncertainty $\Delta \bm{\tau}  \in \mathbb{R}^{12}$. $\Delta \bm{\tau}$ mainly results from internal model mismatches or external disturbances. 

One appealing control strategy in RL framework~\citep{rudin2022learning, lyu2024rl2ac} is to infer a policy network from proprioceptive observations $\bm{o}$ and body commands $\bm{c}$ to desired joint positions $\bm{q}_{d}$, denoted as $\pi(\bm{q}_{d,t}|\bm{o}_t, \bm{c}_t)$. $\bm{o}_t$ includes the base attitude, base angular velocity, joint angles, joint angular velocities, and their historical sequences. Subsequently, a proportional-derivative (PD) controller is employed to track $\bm{q}_{d}$, i.e.,
\begin{align*}
	\bm{\tau} = \bm{K}_p(\bm{q}_{d}-\bm{q}) - \bm{K}_d\bm{\dot{q}}
\end{align*}
with gains $\bm{K}_p \in \mathbb{R}^{{12} \times {12}}$ and $\bm{K}_d \in \mathbb{R}^{{12} \times {12}}$.

The purpose of PhyFilter is to estimate uncertainties $\Delta \bm{\tau}$ using the output of  RL policy and dynamical structure. By defining $\bm{x} = \bm{\dot{q}}$, $\bm{g}(t) = -\bm{C}(\bm{q}, \bm{\dot{q}})\bm{\dot{q}} - \bm{G}(\bm{q}) + \bm{J}^T\bm{F}_c + \bm{\tau}$, and $\bm{f}(t) = \Delta \bm{\tau}$, Eq. \eqref{eq_quadruped_model} can be adjusted to 
\begin{align*}
	\bm{M}(t)\bm{\dot{x}} = \bm{g}(t) + \bm{f}(t).
\end{align*}
Compared with Eq. \eqref{eq_model}, a new term $\bm{M}(t)$ is further induced. Similar to the derivation process from Eq. \eqref{eq_n_filter} to Eq. \eqref{eq_observer}, the PhyFilter algorithm in the locomotion case can be derived as 
\begin{align} \label{eq_observer_quadruped}
	\left\{ {\begin{array}{*{20}{c}}
			\bm{\xi}^{(n)} = -  \bm{g}(t) - \bm{\hat{f}}(t)\\
			\bm{\hat{f}}(t) = a_0\bm{\xi} - \bm{\chi}_1 + \bm{\chi}_2 + a_0 {\int }^{n-1}\left[\int \bm{M}(t) d\bm{x}\right](dt)^{n-1}.
	\end{array}} \right.
\end{align}
where the auxiliary variable $\bm{\xi}$ is defined as $\bm{\xi} = (\bm{\hat f}(t) + \bm{\chi}_1- \bm{\chi}_2)/a_0 - {\int }^{n-1}\left[\int \bm{M}(t) d\bm{x}\right](dt)^{n-1}$. Experiments found that $1$-st order form of PhyFilter (i.e., $n=1$ in Eq. \eqref{eq_observer_quadruped}) is enough to improve the generalization ability (Figs. \ref{quadruped__simulation_result} and \ref{quadruped__experiment_result}).

\textcolor{black}{In practice, the exact inertia matrix $\bm{M}(\bm{q})$ and \textit{Coriolis} force $\bm{C}(\bm{q}, \bm{\dot{q}})$ are difficult to obtain for many complex robots, as accurate analytical modeling becomes increasingly intractable with mechanical complexity. To further probe the tolerance of PhyFilter to such modeling inaccuracy, we deliberately employ simplified $\bm{M}(\bm{q})$ and $\bm{C}(\bm{q}, \bm{\dot{q}})$ in the tests, which can reduce computational complexity. Specifically, $\bm{M}(\bm{q})$ is subject to a diagonalization assumption, where the inertial coupling between joints is neglected and only the diagonal and local inertia terms are retained. $\bm{C}(\bm{q}, \bm{\dot{q}})$ is approximated as linear damping, with the complex velocity-coupled quadratic terms ignored. Despite these simplifications, PhyFilter still significantly improves generalization performance. A formal analysis showing robustness under inaccurate physical priors is provided in the “Robustness analysis under inaccurate priors” section of Supplementary Information.}

\paragraph{Training details}The parallel deep learning training strategy presented in~\citep{rudin2022learning} is adopted here. The key training parameters and domain randomization ranges are listed in Supplementary Table {S1}. The proximal policy optimization (PPO) algorithm~\citep{schulman2017proximal} is used, and the policy with the best training reward is finally adopted to deploy. Note that the training scenario only includes flat terrain with the purpose of highlighting its generalization ability. Moreover, the algorithm is run on a laptop with Intel(R) Core(TM) Ultra and RTX 4060 GPU. For a fair comparison, except for the backend filter mechanism, both the training and testing processes of the proposed algorithm are consistent with those of the baseline.

\subsection{Implementation details of drone flight example}  \label{method_quadrotor}
SEER-I~\citep{jia2025}, a supervised machine learning technique, is employed to learn the mass uncertainty of the drone, which can be formalized as
\begin{align} \label{eq_model_seeri}
	\Delta m \bm{a} = -\bm{\Xi} \bm{\Phi}(\bm{v}_h)\bm{\varsigma}(\Delta m)
\end{align}
with unknown changed mass $\Delta m \in \mathbb{R}^3$, acceleration $\bm{a} \in \mathbb{R}^3$, constant parameter matrix $\bm{\Xi} \in \mathbb{R}^{3 \times (p+1)^4}$, state-related \textit{Chebyshev} polynomial matrix $\bm{\Phi}(\cdot) \in \mathbb{R}^{(p+1)^4 \times (p+1)}$, historical recorded velocity $\bm{v}_h = \{\bm{v}, \bm{v}_1, \cdots, \bm{v}_{n_h}\}$, and mass-related \textit{Chebyshev} polynomial vector $\bm{\varsigma}(\cdot) \in \mathbb{R}^{p+1}$. The model accuracy depends on $p$. $\bm{\Xi}$ can be obtained offline via the least squares (LS) algorithm with labeled data. When applied to an unknown mass scenario, $\bm{\varsigma}(\Delta m)$ can be estimated via a converged adaptive law with known portion $\bm{\Xi} \bm{\Phi}(\bm{v}_h)$. The decomposed structure in Eq. \eqref{eq_model_seeri} can improve the generalization ability subject to unseen mass uncertainty.

In the presence of other kinds of uncertainties (e.g., wind), the above decomposed structure will no longer be suitable, leading to a poor estimation performance (Fig. \ref{dynamics_result_circle_error} (e and f)). Here, we further filter the learning output with physics knowledge, i.e., the dynamics structure. Define $\bm{f}(t) = \Delta m \bm{a}$ with other uncertainties. Similar to the derivation procedure of Eq. \eqref{eq_observer}, one can obtain
\begin{align} \label{eq_observer_quadrotor}
	\left\{ {\begin{array}{*{20}{c}}
			\bm{\xi}^{(n)} = -  mg\bm{e}_z - \bm{\hat{f}}(t)\\
			\bm{\hat{f}}(t) = a_0\bm{\xi} - \bm{\chi}_1 + \bm{\chi}_2 + a_0 {\int}^{n-1}\bm{x}(dt)^{n-1}
	\end{array}} \right.
\end{align}
with nominal mass $m \in \mathbb{R}$, acceleration of gravity $g \in \mathbb{R}$, and its direction $\bm{e}_z\in \mathbb{R}^3$. Moreover, $\bm{f}_{
	\bm{\theta}}$ in $\bm{\chi}_2$ comes from the learning output of SEER-I. The training details are consistent with~\citep{jia2025}. In experiments, it is found that a $1$-st order form of PhyFilter (i.e., $n=1$ in Eq. \eqref{eq_observer_quadrotor}) is enough to improve the generalization ability.

\subsection{Implementation details of aerial manipulation example} \label{method_aerial_manipulator}

The dynamics of the aerial platform, considering the strong coupling effects with a serial manipulator, can be established as:
\begin{align*}
	& m_{b}\dot{\bm{v}}_{b} =  m_{b}g\bm{e}_{z} - f\bm{b}_{3} - \bm{\Delta}_{F} \\
	& \bm{J}\dot{\bm{\omega}}_{b} = -\bm{\omega}_{b}^{\times}\bm{J}\bm{\omega}_{b} + \bm{\tau} + \bm{\Delta}_{\tau}
\end{align*}
where $g\in \mathbb{R}^{+}$ is the gravity and $m_{b}\in \mathbb{R}^{+}$ represents the mass of the aerial platform, $\bm{J}\in \mathbb{R}^{3\times3}$ denotes the inertia matrix of the aerial platform, $f\in \mathbb{R}$ represents the generated force of eight rotors, and $\tau \in \mathbb{R}^{3}$ expresses the generated torque vector. Moreover, $\bm{\Delta}_{F}\in \mathbb{R}^{3}$ and $\bm{\Delta}_{\tau}\in \mathbb{R}^{3}$ are the coupling disturbance force and torque, respectively. $\bm{\omega}_{b}^{\times}$ represents the skew-symmetric operator of $\bm{\omega}_{b}$. According to the centroid principle~\citep{shabana2020dynamics}, the coupling disturbances can be analytically formulated as:
\begin{equation} \label{coupling_V}
	\begin{aligned}  
		\bm{\Delta}_{F} &= -m_{m}\dot{\bm{v}}_{b}+m_{m}g\bm{e}_{z}-m_{m}\bm{R_{b}}\left(\bm{\omega}^{\times}_{b}\left(\bm{\omega}^{\times}_{b}\bm{P}^{B}_{bm}\right) 
		+\dot{\bm{\omega}}^{\times}_{b}\bm{P}^{B}_{bm}+2\bm{\omega}^{\times}_{b}\dot{\bm{P}}^{B}_{bm}+\ddot{\bm{P}}^{B}_{bm}\right) \\ 
		\bm{\Delta}_{\tau} &= -\bm{J^{B}_{mb}}\dot{\bm{\omega}}_{b} - \bm{\omega}^{\times}_{b}\bm{J^{B}_{mb}}\bm{\omega}_{b}+m_{m}P^{B}_{bm}\bm{R_{b}}^{\top}\left(g\bm{e}_{z}-\dot{\bm{v}}_{b}\right)
		- \bm{\dot{J}^{B}_{mb}}\dot{\bm{\omega}}_{b} - \bm{\omega}^{\times}_{b}\bm{L}^{B}_{m} - \dot{\bm{L}}^{B}_{m}
	\end{aligned}
\end{equation}
where $m_{m}\in \mathbb{R}^{+}$ is the mass of the manipulator, $\bm{P}^{B}_{bm}\in \mathbb{R}^{3}$ is the center of mass of the manipulator, $\bm{J^{B}_{mb}}\in \mathbb{R}^{3\times3}$ is the inertia tensor of the manipulator along the body frame, and $\bm{L}^{B}_{m}\in \mathbb{R}^{3}$ describes the angular momentum of the manipulator. It can be proven that the unknown high-order term is an analytic function regarding the state and inertial parameters of the aerial manipulator. The detailed modelling analysis of the coupling disturbances can be found in~\citep{10339889}.

Let $\eta\in\mathbb{R}^{3}$ be the system state and $\varphi \in \mathbb{R}^{n_{\varphi}}$ be the unknown parameter vector. On the basis of decomposition theorem~\citep{2024arXiv240713229J}, $\Delta_{F}(\eta,\varphi)$ can be decomposed with arbitrary accuracy as:
\begin{equation*}
	\bm{\Delta}_{F}(\eta,\varphi) = \bm{\Xi_{p}}\bm{\Phi_{p}}(\eta)\bm{\xi}_{p}(\varphi) + \mathcal{O}(n) 
\end{equation*}
where $\bm{\Xi_{p}} \in \mathbb{R}^{3\times(p+1)^{3+n_{\varphi}}}$ is an unknown parameter weight matrix to be learned.  $\bm{\Phi_{p}}(\eta)$ and $\bm{\xi}_{p}(\varphi)$ are mappings with corresponding dimensions. $p\in\mathbb{R}$ is the hyperparameter determining the decomposition accuracy. 
In the presence of additional uncertainties (e.g., model uncertainties and external wind), the aforementioned decomposed structure is no longer suitable, resulting in a poor estimation performance. Hence, the learning output is further filtered via physics knowledge, i.e., the dynamic structure. The filter framework can be expressed in the same form as Eq. \eqref{eq_observer_quadrotor} by treating $\bm{f}(t) = \bm{\Delta}_{F}(\eta,\varphi)$ with other uncertainties.

Similarly, the coupling torque can be decomposed as:
\begin{equation*}
	\bm{\Delta}_{\tau}(\eta,\vartheta) = \bm{\Xi_{\tau}}\bm{\Phi_{\tau}}(\eta)\bm{\xi}_{\tau}(\vartheta) + \mathcal{O}(n)
\end{equation*}
where $\bm{\Xi_{\tau}} \in \mathbb{R}^{3\times(p+1)^{3+n_{\vartheta}}}$ is the learned parameter weighting matrix.  $\bm{\Phi_{\tau}}(\eta)$ and $\bm{\xi}_{\tau}(\vartheta)$ are mappings with corresponding dimensions, which can consist of the known Chebyshev polynomials. The filter framework for the rotational loop can be derived similarly to Eq. \eqref{eq_observer_quadrotor}. 

\paragraph{Training details} 
During the process of collecting training data, the manipulator is commanded to follow a circular trajectory designed to excite the nonlinear coupling dynamics continuously. The reference trajectory is defined as $\left[0.28+{0.1\sin(\omega t)}/{\sqrt{5}},{0.1\sin(\omega t)}/{\sqrt{5/4}},0.105+0.1\cos(\omega t)\right]$ m in its base frame, which forms a three-dimensional periodic path covering both vertical and horizontal planes. To further enrich the dynamic excitation, the angular frequency of the trajectory is gradually increased from 2.5 to 4.5 rad/s. It is ensured that all joint dynamics of the manipulator are sufficiently excited. Moreover, Eq. \eqref{coupling_V} implies that the nonlinear coupling effects are associated with these signals $\dot{\bm{v}}_{b}$, $\dot{\bm{\omega}}_{b}$, $\dot{\bm{P}}^{b}_{bm}$, and $\ddot{\bm{P}}^{b}_{bm}$ that cannot be accurately measured by onboard sensing. Therefore, the time-delay embedding theorem~\citep{takens2006detecting, sauer1991embedology} is used in training, that is, the historical records of signal can reflect its differentiation. Concretely, the collected dataset includes linear velocity, attitude, angular velocity, joint position, and joint velocity, along with three historical values of each signal to encode their derivative information. Other training settings follow those used in the preceding drone example.

\subsection{Implementation details of perception example} \label{method_accelerator}

In the absence of a high-performance accelerometer for drones, the acceleration can be obtained through historical velocities, i.e.,
\begin{align*}
	\bm{a} = \bm{f}(\bm{v}_h)
\end{align*}
with acceleration $\bm{a} \in \mathbb{R}^3$, historical recorded velocity $\bm{v}_h = \{\bm{v}, \bm{v}_1, \cdots, \bm{v}_{n_h}\} \in \mathbb{R}^{3n_h}$, and a mapping $\bm{f}(\cdot)$. For example, $\bm{f}(\cdot)$ can be designed as the discrete differentiator (DD), tracking differentiator (TD)~\citep{han2009pid}, robust differentiator (RD)~\citep{seeber2021robust} with a predefined convergence time, or neural network-based accelerator~\citep{9635983}.

To further enhance the generalization of the neural network-based accelerator, we filter its output using the underlying kinematic structure, i.e.,
\begin{align*} 
	\left\{ {\begin{array}{*{20}{c}}
			\bm{\xi}^{(n)} = - \bm{\hat{f}}(t)\\
			\bm{\hat{f}}(t) = a_0\bm{\xi} - \bm{\chi}_1 + \bm{\chi}_2 + a_0 {\int}^{n-1}\bm{x}(dt)^{n-1}
	\end{array}} \right.
\end{align*}
with $\bm{f}_{
	\bm{\theta}}$ in $\bm{\chi}_2$ comes from the learning output of the learned neural network. The ablation study on different orders $n$ and parameters $\bm{a}_f$ is conducted, as shown in Fig. \ref{perception_result0}e. The parameters $\bm{a}_f$ can be set according to the \textit{pole-placement} theory or the proposed auto-learned algorithm (Algorithm 2).

\paragraph{Training details} The employed neural network has two hidden layers with $20$ hidden units each. The \textit{trainscg} optimizer is used. In Fig. \ref{perception_result0} (c-e), the training datasets consist of $28000$ samples within $0\sim 2\ m/s$, while the testing datasets consist of $8400$ samples within $0\sim 3\ m/s$. All compared methods are tuned to their best performance~\citep{10288520}.

\bibliographystyle{assets/plainnat}
\bibliography{paper}

\clearpage
\newpage
\appendix

	\renewcommand*{\thefigure}{S1}
\begin{figure*}
	\centering
	\includegraphics[width=1\linewidth]{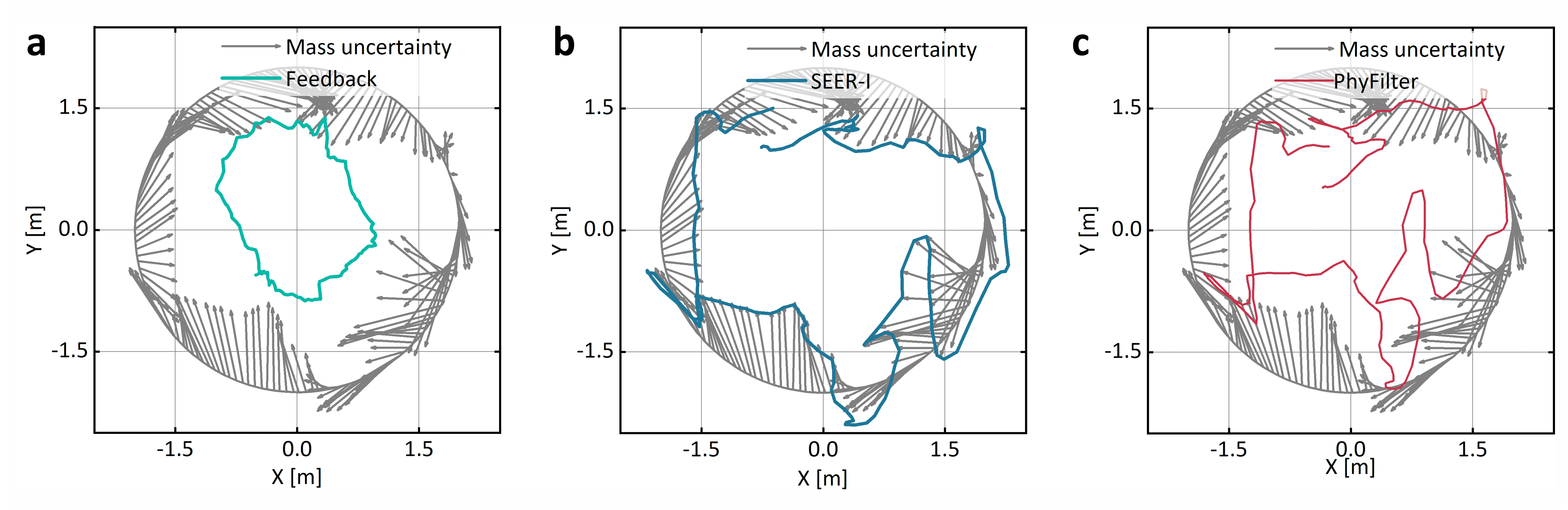}
	\caption{\textbf{Estimation performance of the drone example.} \textbf{a}, Uncertainty estimation with feedback-based disturbance observer. \textbf{b}, Uncertainty estimation with SEER-I. \textbf{c}, Uncertainty estimation with the proposed PhyFilter. For reference, the ground truth of mass uncertainty is provided in (\textbf{a})-(\textbf{c}), which can be deduced from the ground-truth mass in advance. It can be seen that SEER-I can only reflect the mass uncertainty, while PhyFilter can further capture other unknown wind and inherent uncertainties, leading to more accurate trajectory tracking, as shown in Fig. 4e.}
	\label{dynamics_result_circle_uncer}
\end{figure*}

\renewcommand*{\thefigure}{S2}
\begin{figure*}
	\centering
	\includegraphics[width=0.9\linewidth]{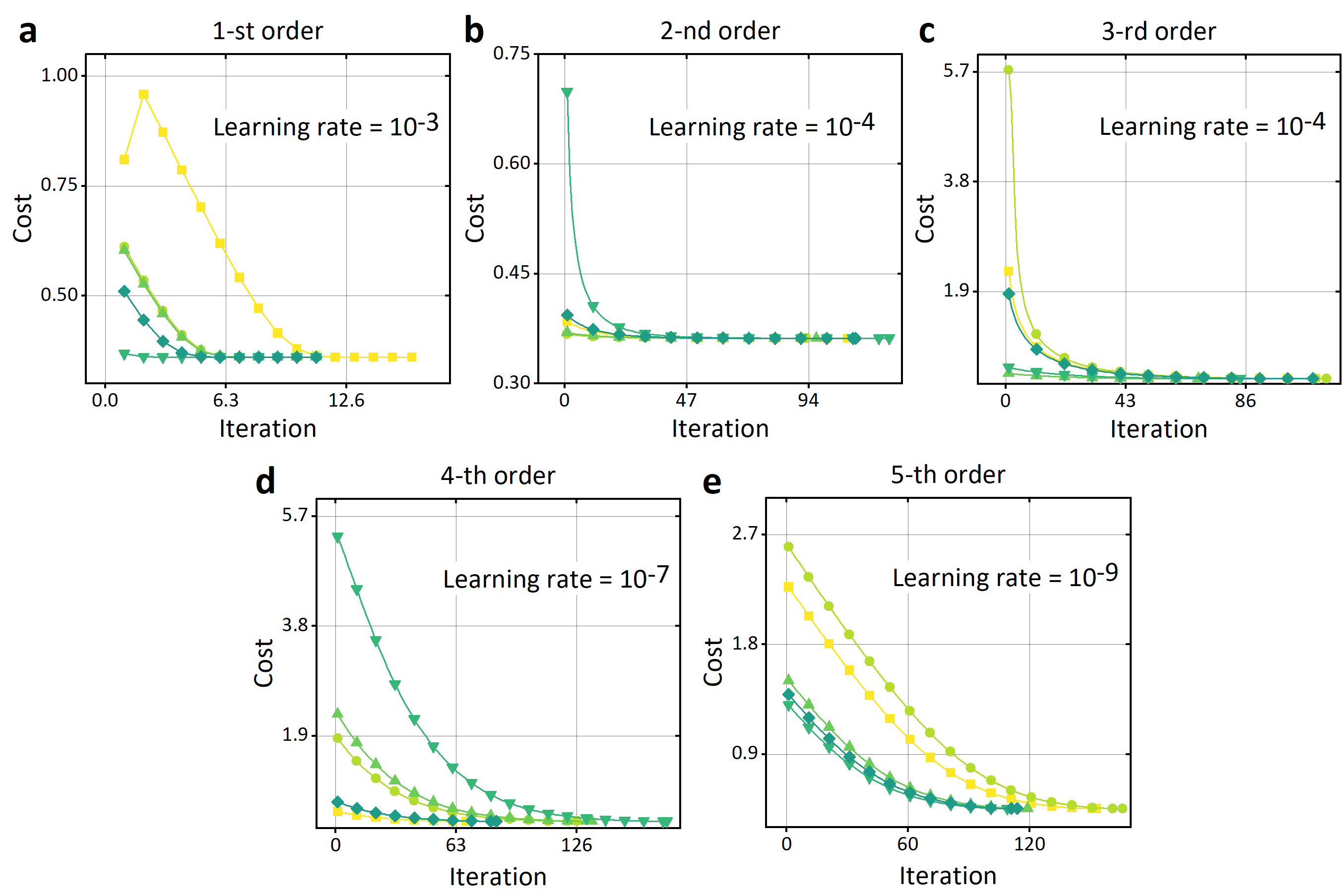}
	\caption{\textbf{The convergence performance of Algorithm 2 at different filter orders in the acceleration example.} Each case is repeated $5$ times with different initial filter parameter $\bm{a}_{f0}$. A higher filter order requires a smaller learning rate.}
	\label{learning_convergence}
\end{figure*}

\renewcommand*{\thefigure}{S3}
\begin{figure*}
	\centering
	\includegraphics[width=0.5\linewidth]{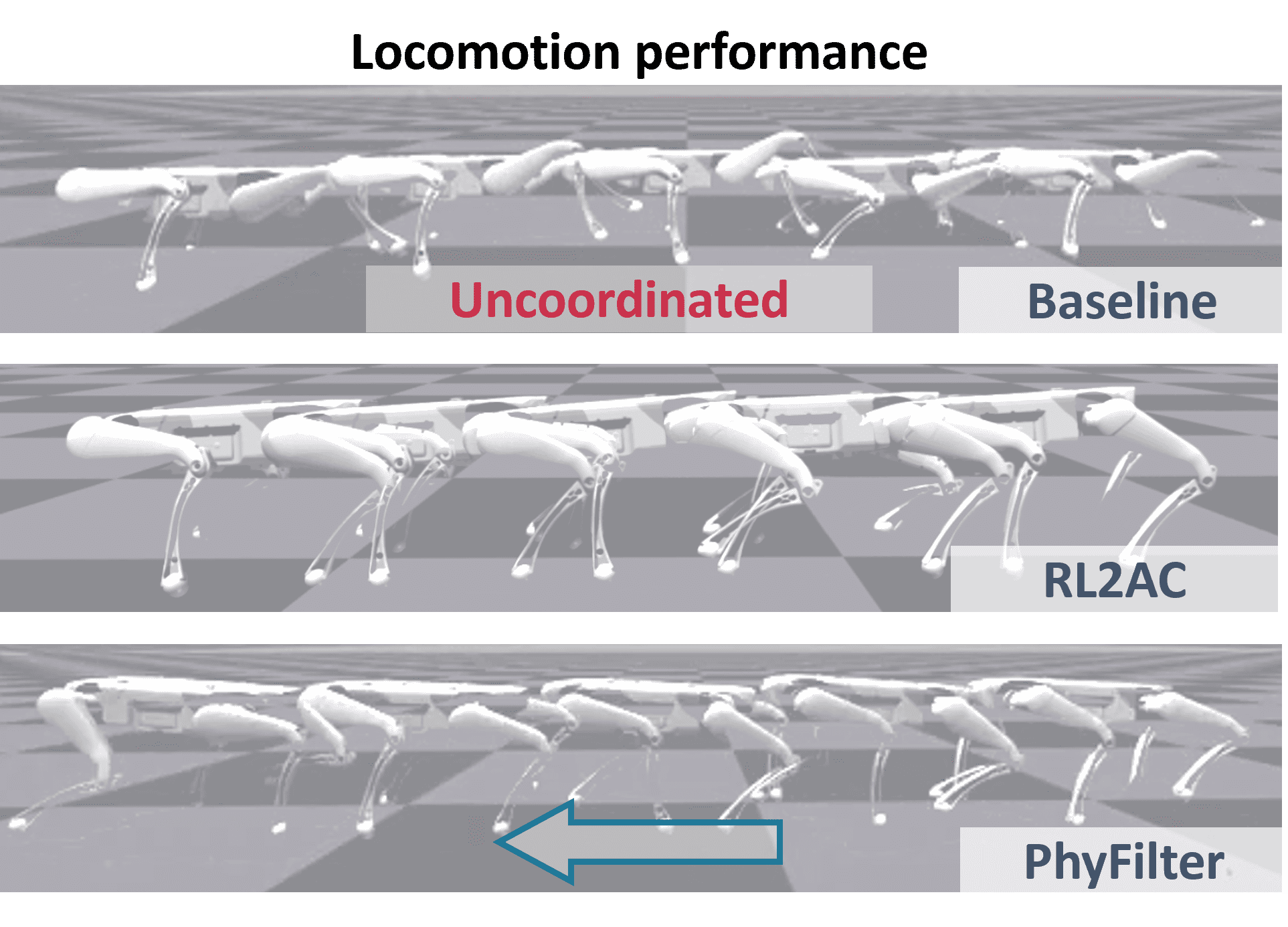}
	\caption{\textbf{The trained locomotion performance of all compared methods.} PhyFilter and RL2AC can yield more coordinated locomotions, whereas the baseline results in inferior performance.}
	\label{locomotion_performance}
\end{figure*}

\renewcommand*{\thefigure}{S4}
\begin{figure*}
	\centering
	\includegraphics[width=1\linewidth]{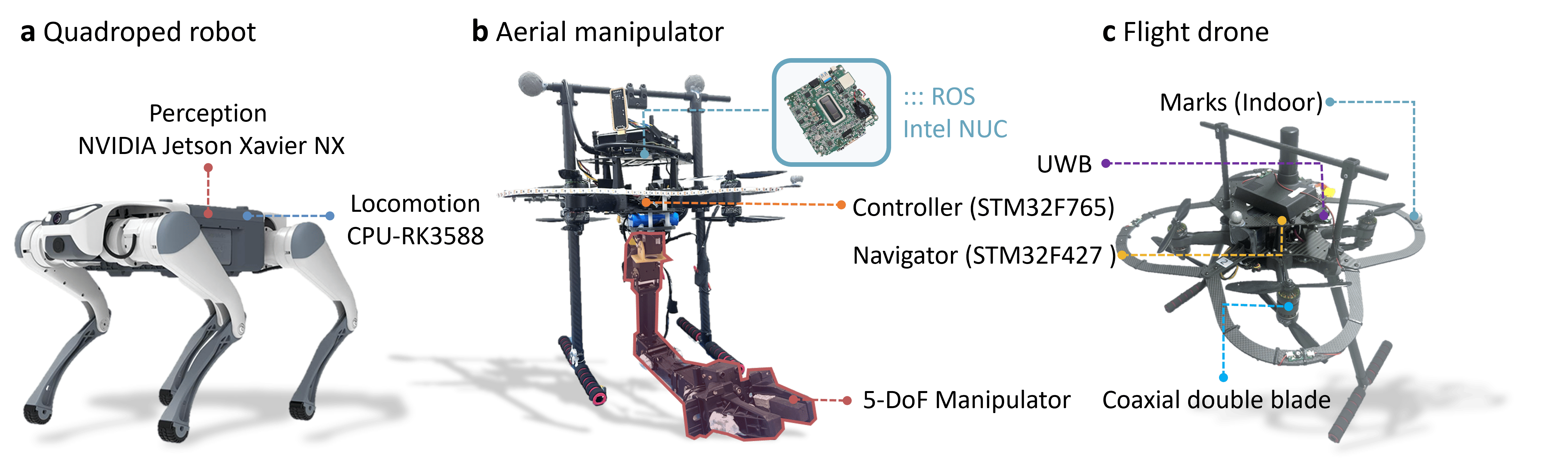}
	\caption{\textbf{The employed hardwares.} \textbf{a}, The quadruped robot used in the locomotion example. \textbf{b}, The aerial manipulator used in the manipulation example. \textbf{c}, The flight drone used in the flight example.}
	\label{hardware}
\end{figure*}

\renewcommand*{\thefigure}{S5}
\begin{figure*}
	\centering
	\includegraphics[width=1\linewidth]{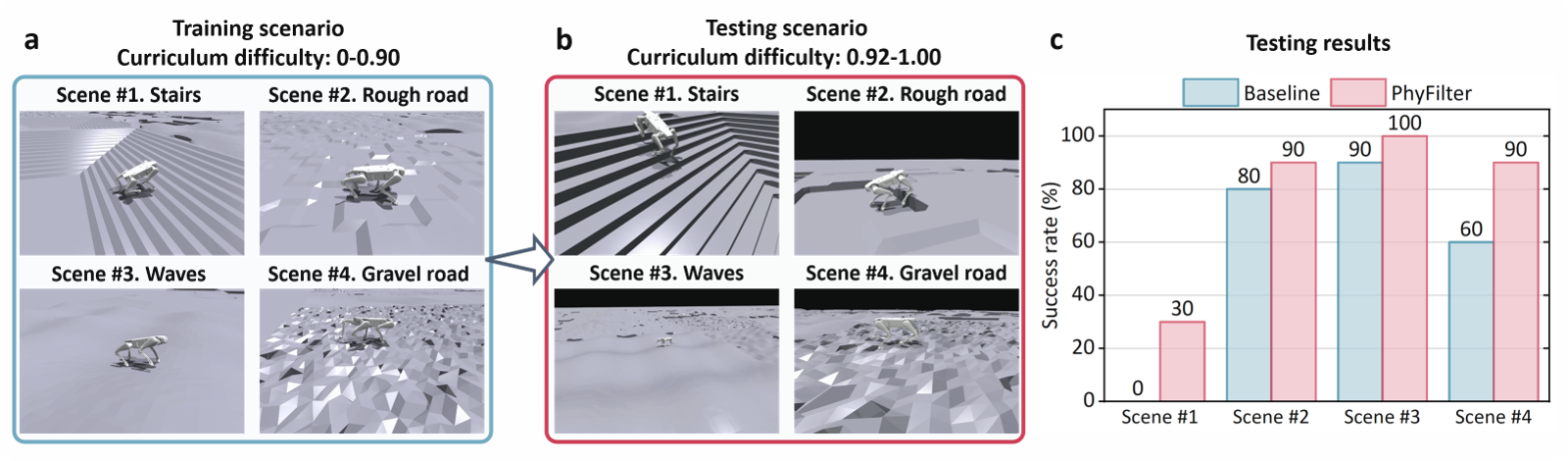}
	\caption{\textcolor{black}{\textbf{Comparison between the baseline and PhyFilter under a strongly randomized training setting.} \textbf{a}, Training scenarios at curriculum difficulty 0–0.90 (Scenes \#1–\#4: stairs, rough road, waves, gravel road). \textbf{b}, Testing scenarios at curriculum difficulty 0.92–1.00, the same terrains but beyond the training distribution. \textbf{c}, Testing success rates (\%) across the four scenes. Even trained with full standard randomization, the baseline still fails under sufficiently out-of-distribution conditions, whereas PhyFilter consistently achieves higher success rates.}}
	\label{full_dr}
\end{figure*}

\renewcommand*{\thefigure}{S6}
\begin{figure*}
	\centering
	\includegraphics[width=1\linewidth]{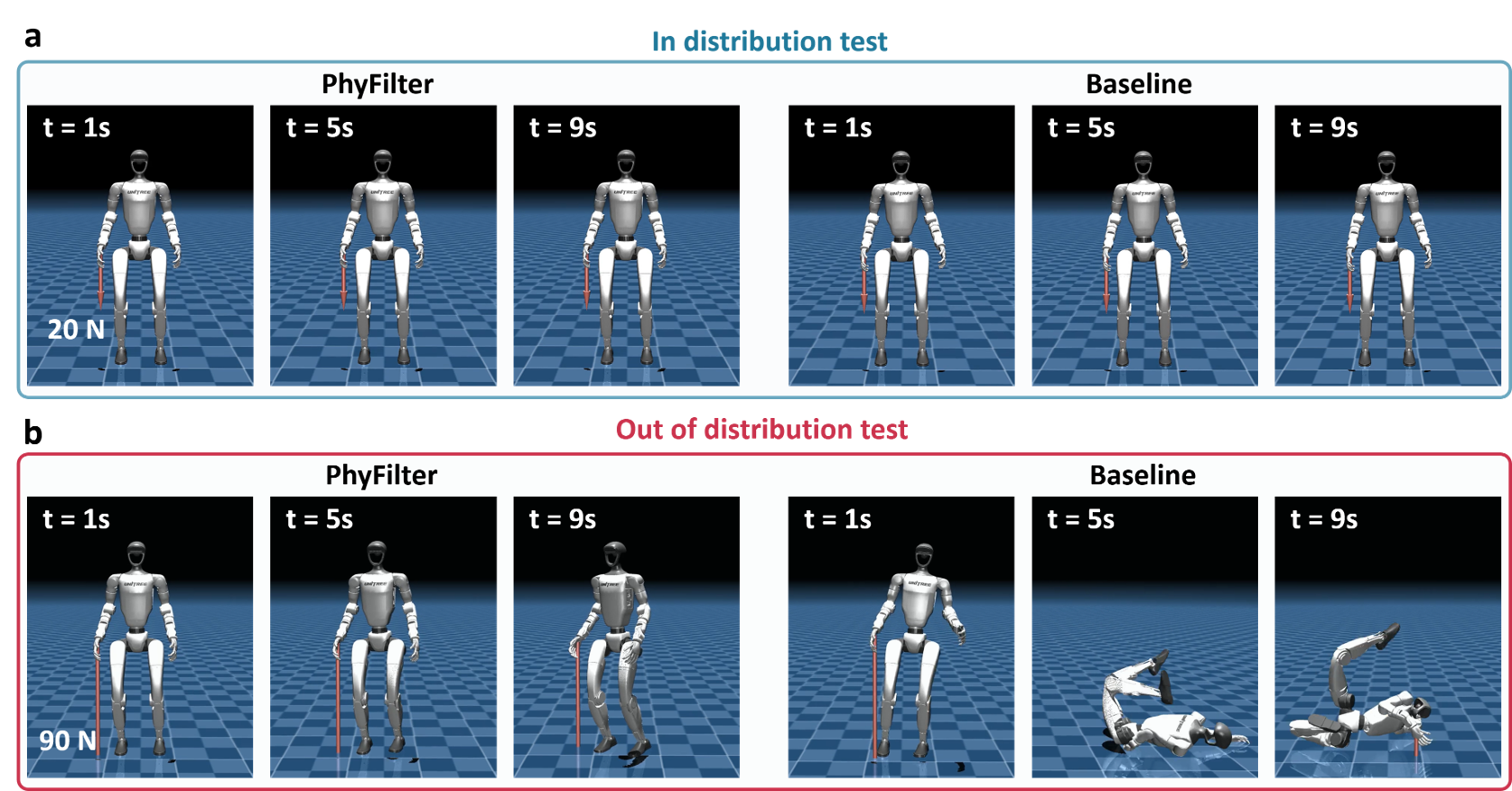}
	\caption{\textcolor{black}{\textbf{Humanoid robot tests under in-distribution and out-of-distribution external forces.} A lateral force is applied to one arm of the humanoid, and each row shows snapshots of PhyFilter (left) and the baseline (right) at t=1, 5, and 9s. \textbf{a}, In-distribution test under a 20 N force: both PhyFilter and the baseline keep the robot stably standing. \textbf{b}, Out-of-distribution test under a 90~N force: PhyFilter maintains stable standing throughout, whereas the baseline loses balance and falls.}}
	\label{appen_humanoid}
\end{figure*}

\clearpage

\renewcommand*{\thetable}{S1}
\begin{table}
	\centering
	\begin{tabular}{cc|cc}
		\hline
		\hline
		\textbf{Training parameter} & \textbf{Value} & \textbf{Domain randomization} & \textbf{Range} \\
		\hline
		\hline
		Number of environments & $4096$ &	Mass & $-1\sim 3$ kg \\
		\hline
		Number of iterations & $15000$& Velocity & $-1 \sim 1$ m/s \\
		\hline
		Filter order $n$  & $1$ & Gain $\bm{K}_p$ & $0.8\sim 1.2$ \\
		\hline
		Filter parameter $a_0$ & $1$ & Gain $\bm{K}_d$ & $0.8\sim 1.2$ \\ 
		\hline
		&  & Friction & $0.1\sim 1.25$ \\
		\hline
		\hline
	\end{tabular}
	\caption{Training parameters and domain randomization ranges for the quadruped example.}
	\label{RL_param}
\end{table}

\clearpage

\section{Related work} \label{sec_related}

\subsection{Policy learning for quadruped robots}

Benefiting from cheaper yet more efficient hardware (e.g., proprioceptive actuators~\citep{wensing2017proprioceptive}), advanced control algorithms around optimal control and RL, and high-fidelity simulators (e.g., MuJoCo~\citep{todorov2012mujoco}, Isaac Gym~\citep{makoviychuk2021isaac}), quadruped robots are currently seeing resurgent interest~\citep{ha2024learning}. In particular, RL is currently the most mainstream algorithm with various implementation manners, like imitation learning~\citep{han2024lifelike}, hierarchical learning~\citep{han2024lifelike, yang2020multi}, privileged learning~\citep{kumar2021rma, lee2024learning}, and competitive learning~\citep{kim2025high}. Despite the fact that adding noise during training can improve the robustness, there is still a major sim-to-real gap in scenarios that exceed the training distribution/training scenarios~\citep{shi2024rethinking}. 

Nowadays, plenty of researchers attempt to incorporate classical control schemes and other \textit{a priori} physical knowledge into RL to improve the learning generalization~\citep{10286076}. For example, ~\citep{lyu2024rl2ac} discovers that the contact force of the quadruped robot under an RL-based PD controller can be approximated by a linearly parameterized form with respect to the joint tracking error. Hence, a composite adaptive controller can be naturally employed to estimate the unseen disturbances, narrowing the sim-to-real gap. In~\citep{humphreys2024learning}, a bio-inspired gait scheduler is incorporated into RL to encode pseudo gait procedural memory and adaptive motion adjustment, leading to an improved generalized performance for quadruped locomotion. ~\citep{jenelten2024dtc} employs a model-based planner to guide the learning policy, so that the sparse reward problem of RL can be avoided while maintaining robustness. Our presented PhyFilter also attempts to incorporate a control feedback scheme into RL, yielding a more lightweight and interpretable framework.

\subsection{Dynamics learning for multirotor drones}

Achieving precise maneuver control of drones requires effective handling of external disturbances and internal model uncertainties. This part mainly reviews the learning-dominant methods. ~\citep{kaufmann2023champion} designs an end-to-end control strategy for the quadrotor by employing RL, which can achieve the level of human world champions in drone competitions. To address uncertainties, this work trains the control strategy using the domain randomization~\citep{tobin2018domain} and constructs an empirical noise model to reduce the difference between simulation and reality. ~\citep{10611665} employs residual networks to estimate dynamical uncertainties explicitly, which is then regarded as the learning input of an end-to-end RL. ~\citep{song2023reaching} compares the performance of RL and optimal control in drone competitions. This work finds that RL has a higher success rate and maneuvers faster in the presence of uncertainties. A neural network-based estimation algorithm is proposed in~\citep{zhaonature} to predict the downwash-induced disturbance forces and torques acting on vertically stacked drones. The results demonstrate that incorporating \textit{Gaussian}-predicted airflow velocities as input features significantly enhances the estimation accuracy.

The approach of integrating traditional control and advanced learning for the multirotor drone control is increasingly appealing~\citep{saviolo2023learning}. $\mathcal{L}_1$ adaptive observer-enhanced RL is presented in~\citep{huang2023datt}, enabling the drone to precisely control under multiple disturbances. Whereas, the estimation error of $\mathcal{L}_1$ adaptive observer is not considered in RL training. The \textit{Koopman} operator is employed in~\citep{10288520} to learn the uncertainty model online, favoring the convergence of a model-based uncertainty observer. It can be shown that the \textit{Koopman} operator-based observer constitutes a special case of PhyFilter.

\subsection{Precise control of aerial manipulators}

Different from bare multirotors, achieving precise control of an aerial manipulator presents substantial challenges~\citep{9476743,9462539,SR_PHRI}. Firstly, the control accuracy of the aerial platform is highly susceptible to model uncertainties. In practice, the assumption that a three-dimensional model cannot perfectly match reality due to the inevitable presence of motors, sensors, and other components~\citep{9919418,10547187,HeG-RSS-25}. In addition, the dynamics of the aerial platform and the manipulator are intrinsically coupled, as the multi-link manipulator is rigidly mounted on the drone. As pointed out in~\citep{9309359,10339889}, the strong dynamic coupling disturbances are significantly influenced by the relative motion of the manipulator with respect to the aerial platform. The strong coupling effects further exacerbate the instability of the aerial platform. The reason is that explosive dynamic coupling terms affect its dynamics. Beyond this, external disturbances like aerodynamic effects play a critical role in degrading the dynamic performance of the aerial platform~\citep{10339889,11025148,10935295,10994382}.

A common approach to address coupling disturbances is the use of disturbance observers. For instance, an adaptive sliding-mode disturbance observer with a prescribed convergence-time bound is proposed in~\citep{9768118} to estimate state-dependent uncertainties and external disturbances. In~\citep{10994382}, an acceleration-based estimation method is presented to estimate total external wrench including modeling errors and wind disturbances. However, when external disturbances are coupled with system states, the assumption of bounded uncertainty derivatives becomes impractical. This is primarily because one cannot assume that the states are bounded before the closed-loop stability of the system is established~\citep{2024arXiv240713229J,10145100}. In~\citep{11073186}, a characteristic modeling with the bounded identification error is introduced into the integral controller to guarantee the finite-time stability of the aerial manipulator. Furthermore, in~\citep{10083097}, two extended state observers are incorporated into a cascade control scheme to improve quadrotor performance by accounting for first-order coupling effects. Nevertheless, these ESOs rely on a uniform regulating mechanism to adjust different observed states. Such a framework may lead to inappropriate parameter tuning, thereby degrading the dynamic 
performance of the observers. Although observer-based estimation techniques have yielded promising results, the prior knowledge of the system dynamics is not fully exploited to enhance estimation accuracy~\citep{9768118,10994382,2024arXiv240713229J,10083097,10145100}.

Driven by the availability of large-scale datasets and advances in modern learning algorithms, data-driven learning has made remarkable progress. Preliminary efforts have been devoted to disturbance learning. For example, a radial basis function neural network with an adaptive weight update law has been developed to compensate for lumped disturbances, including model uncertainties and coupling effects, in real time~\citep{CAO2022107607,10547187}. In~\citep{CAI2025106418}, an adaptive approach is presented to compensate for the unmodeled nonlinear disturbances brought by the manipulator motion. Moreover, a gaussian process regression is developed to estimate the external environment disturbances~\citep{10935295}. The well-known generalization problem arises when the system encounters scenarios beyond the training dataset~\citep{jia2024feedback,nerual_fly}. Although offline training algorithms can be used to initialize neuron weights~\citep{10547187,10706036}, the inevitable online adaptation required to cope with time-varying disturbances often results in sub-optimal transient responses.

\subsection{Online differentiator}
Calculating the precise derivative of a measured signal in real time is a crucial topic with numerous practical uses, including reliable state estimation. The most straightforward analytical differentiator is the finite difference-based discrete differentiator (DD), which approximates the derivative of discrete-time signals using differences between their sampled values. As a core component of active disturbance rejection control, the tracking differentiator (TD)~\citep{han2009pid} performs differentiation on noisy signals through a specially designed \textit{sign} function. ~\citep{seeber2021robust} propose a robust differentiator (RD) capable of converging to the derivative of a given signal within a predefined time frame. Nevertheless, these aforementioned analytical algorithms involve cumbersome parameter tuning procedures, and variations in signal characteristics may necessitate corresponding parameter adjustments. Learning-based methodologies represent another effective approach for deriving the derivative of measured signals. For instance, ~\citep{9635983} formulates a neural acceleration estimator to achieve accurate prediction of the motion of uneven free-flying objects. However, as with other neural network-based application scenarios, the generalization problem is an intractable difficulty that cannot be avoided.

\subsection{Physics-informed machine learning}

Physics-informed machine learning~\citep{karniadakis2021physics, tobin2017domain, tobin2018domain, chen2018neural, jia2024feedback, raissi2019physics, wang2024imperative, greydanus2019hamiltonian, cranmer2020lagrangian} is designed to integrate prior physical insights into the development of machine learning methods. The primary objectives of this integration include enhancing model interpretability, accelerating training convergence, boosting predictive precision, and strengthening generalization capabilities. Three synergistic strategies are commonly adopted. These involve introducing observational~\citep{tobin2017domain, tobin2018domain}, inductive~\citep{chen2018neural, jia2024feedback}, and learning~\citep{raissi2019physics, wang2024imperative, greydanus2019hamiltonian, cranmer2020lagrangian} biases into the training dataset, learning structure, and optimization algorithm, respectively. For instance, ~\citep{lutter2023combining} demonstrates that neural networks can be trained to directly approximate the mass matrix and potential energy functions derived from \textit{Lagrangian} and \textit{Hamiltonian} mechanics. Such physics-informed structures yield more physically consistent models compared to conventional black-box neural network architectures. Along a parallel research trajectory focused on neural ODEs~\citep{chen2018neural}, ~\citep{greydanus2019hamiltonian} propose the \textit{Hamiltonian} neural network, which is specifically engineered to learn the \textit{Hamiltonian} of dynamic systems. A key advantage of this approach is its ability to preserve the energy conservation property inherent to such systems. However, a notable constraint of the \textit{Hamiltonian} neural network is its reliance on canonical coordinates. To address this limitation, ~\citep{cranmer2020lagrangian} develops the \textit{Lagrangian} neural network, which eliminates the need for canonical coordinates.

In the domain of learning-based robotic autonomy, where improving generalization remains a critical challenge, ~\citep{wang2024imperative} introduces a self-supervised neural-symbolic learning framework termed imperative learning. This framework formulates the learning task as a bilevel optimization problem, wherein the base learner leverages symbolic knowledge, encompassing physical principles and logical reasoning, to enhance performance. ~\citep{romero2024actor} explores the integration of model predictive control (MPC) with actor-critic RL. Their findings indicate that this hybrid approach enables the simultaneous retention of MPC's short-term optimization strengths and the end-to-end training benefits characteristic of RL. A physics-informed world model is developed in~\citep{li2025pin} for bib-prehensile manipulation, in which a physics-aware randomization strategy is employed to bridge the sim-to-real gap. Inspired by a physics-based aerodynamic model, ~\citep{zhaonature} incorporates airflow velocities as feature inputs to the neural network, enabling precise estimation of disturbance force and torque for vertically stacked drones.

\subsection{Generalization improvement}

\subsubsection{Domain randomization}
To enable data scaling, various simulational physics engines~\citep{todorov2012mujoco, makoviychuk2021isaac, geng2025roboverse} have been established to train robotic agents. The utilized data is synthesized from massively different settings by randomizing the parameters of the simulator, also known as domain randomization~\citep{tobin2017domain, tobin2018domain}, aiming to bridge the sim-to-real gap. Several drawbacks exist in the paradigm. At first, domain randomization pursues the average performance among randomized environments, i.e., sacrificing accuracy for robustness~\citep{jia2024feedback, ha2024learning, green2012linear}, similar
to classical robust control~\citep{ha2024learning, green2012linear}. Secondly, substantial computing resources are often required to train such a monolithic model, which may be infeasible for compute-constrained cases. Finally, universal physics engines are incapable of encompassing the full spectrum of diverse real physical robots and their dynamical environments. The standardization of data format, software, and hardware still remains a long-term endeavor~\citep{wu2024robomind}.

\subsubsection{Domain adaptation}
Apart from domain randomization, domain adaptation~\citep{kouw2019review} is another way to combat the generalization problem, which tries to identify the encountered environment explicitly, so that current decisions can be appropriately adjusted~\citep{kouw2019review}. This scheme is akin to adaptive control~\citep{ha2024learning, slotine1991applied}. In comparison with domain randomization, while domain adaptation is trained more complexly, it maintains accuracy in specific scenarios. A particular online domain adaptation case, test-time adaptation~\citep{kouw2019review}, aims at updating the pre-trained model only through unlabeled test data, considering the source data is usually inaccessible during online testing~\citep{chidlovskii2016domain}. However, domain adaptation also relies on substantial computing resources to train a monolithic model.

\subsubsection{Filtering learning output}
Few previous works use physics knowledge to filter the outcome of machine learning. A differentiable \textit{Kalman} filter augmented neural acceleration estimator is designed in~\citep{9635983, yin2021rall}, in which the \textit{Kalman} filer balances the measurement (i.e., learning output) and prediction from \textit{a priori} propagation model. However, the targeted uncertainty of the above method is limited to random white noise. 

\section{Formalization of parameter learning} \label{sec_learning_appen}

The filter parameter $\bm{a}_f = \{a_0, a_1, \cdots, a_{n-1}\}$ in Eq. (\textcolor{blue}{3}) can be set according to the mature \textit{pole-placement} theory. To avoid the domain-specific expertise required, an auto-learning algorithm is further developed to learn the filter parameter $\bm{a}_f$ from the aspect of an optimal control scheme.  

Before the procedure, define $\bm{\bar{\xi}} = \left[\bm{\xi}^T, \bm{\dot{\xi}}^{T}, \cdots, \bm{\xi}^{(n-1)T}\right]^T \in \mathbb{R}^{n_xn}$, $\bm{\alpha}_i ={\int }^{i}\bm{\hat {f}}(t)(dt)^i \in \mathbb{R}^{n_x}$, $\bm{\bar{\alpha}}=\left[\bm{\alpha}_1^T, \bm{\alpha}_2^{T}, \cdots, \bm{\alpha}_{n-1}^{T}\right]^T\in \mathbb{R}^{n_x(n-1)}$, $\bm{\beta}_i ={\int }^{i}\bm{{f}}_{\bm{\theta}}(t)(dt)^i$, and $\bm{\bar{\beta}} = \left[\bm{\beta}_1^T, \bm{\beta}_2^{T}, \cdots, \bm{\beta}_{n-1}^{T}\right]^T\in \mathbb{R}^{n_x(n-1)}$. Rearrange Eq. (\textcolor{blue}{5}), leading to
\begin{align} \label{eq_opti_obj}
	\bm{\hat{f}}(\bm{\bar{\xi}}, \bm{\bar{\alpha}}, \bm{\bar{\beta}}, \bm{a}_f) = a_0 \bm{{\xi}} + \bm{f}_{\bm{\theta}}(\bm{x}, \bm{z}) + a_0{\int}^{n-1}\bm{x}(dt)^{n-1} - \sum\limits_{i = 1}^{n - 1} {{a_{n - i}}{\bm{\alpha} _i}}  + \sum\limits_{i = 1}^{n - 1} {{a_{n - i}}{\bm{\beta} _i}}
\end{align}
where
\begin{align} \label{eq_opti_cons}
	\bm{\dot{\bar{\xi}}} & =  \bm{\phi}_{\bm{\bar{\xi}}}(\bm{\bar{\xi}}, \bm{\bar{\alpha}}, \bm{\bar{\beta}}, \bm{a}_f) = \left[\bm{\dot{\xi}}^{T}, \bm{\ddot{\xi}}^{T}, \cdots, \bm{{\xi}}^{(n-1)T}, (-\bm{g}-\bm{\hat{f}})^T\right]^T \notag \\
	\bm{\dot{\bar{\alpha}}} &= \bm{\phi}_{\bm{\bar{\alpha}}}(\bm{\bar{\xi}}, \bm{\bar{\alpha}}, \bm{\bar{\beta}}, \bm{a}_f) = \left[\bm{\hat{f}}^T, \bm{\alpha}_1^T, \cdots, \bm{\alpha}_{n-3}^T, \bm{\alpha}_{n-2}^T \right]^T \notag \\
	\bm{\dot{\bar{\beta}}} &= \bm{\phi}_{\bm{\bar{\beta}}}(\bm{\bar{\beta}}, \bm{a}_f)= \left[\bm{{f}}_{\bm{\theta}}^T, \bm{\beta}_1^T, \cdots, \bm{\beta}_{n-3}^T, \bm{\beta}_{n-2}^T \right]^T.
\end{align}

Regard the above equations Eqs. \eqref{eq_opti_obj}-\eqref{eq_opti_cons} as an optimal control system with concatenated states $\{\bm{{\bar{\xi}}}, \bm{{\bar{\alpha}}}, \bm{{\bar{\beta}}}\}$ and control input $\bm{a}_f$. Then, the optimization problem to find optimal $\bm{a}_f$ can be formalized as 
\begin{align*}
	\mathop {\min }\limits_{\bm{a}_f} J(\bm{a}_f) & = \mathop {\min }\limits_{\bm{a}_f} \sum\limits_{i = 1}^{N - 1}l_i(\bm{f}_i^*, \bm{\hat{f}}_i, \bm{a}_f) + l_N(\bm{f}_i^*, \bm{\hat{f}}_i) \notag \\
	s.t.\quad \bm{{\bar{\xi}}}_{i+1} & = \bm{\phi}_{\bm{{\bar{\xi}}}}^d(\bm{\bar{\xi}}_i, \bm{\bar{\alpha}}_i, \bm{\bar{\beta}}_i, \bm{a}_f), \quad i = {1,2,\cdots,N-1}  \notag \\
	\bm{{\bar{\alpha}}}_{i+1} &= \bm{\phi}^d_{\bm{\bar{\alpha}}}(\bm{\bar{\xi}}_i, \bm{\bar{\alpha}}_i, \bm{\bar{\beta}}_i, \bm{a}_f), \quad i = {1,2,\cdots,N-1} \notag \\
	\bm{{\bar{\beta}}}_{i+1} &= \bm{\phi}^d_{\bm{\bar{\beta}}}(\bm{\bar{\beta}}_i, \bm{a}_f), \quad i = {1,2,\cdots,N-1}
\end{align*}
where $N$ denotes the sample number, $l_i(\cdot) \in \mathbb{R}$ is defined to quantify the
state differences between model rollout $\bm{\hat{f}}_i$ and labelled sample $\bm{{f}}_i^*$, $l_N(\cdot) \in \mathbb{R}$ represents the terminal cost, $\bm{\phi}_{\bm{{\bar{\xi}}}}^d(\cdot)$, $\bm{\phi}_{\bm{{\bar{\alpha}}}}^d(\cdot)$, and $\bm{\phi}_{\bm{{\bar{\beta}}}}^d(\cdot)$ refer to the discretized integration of $\bm{\phi}_{\bm{{\bar{\xi}}}}(\cdot)$, $\bm{\phi}_{\bm{{\bar{\alpha}}}}(\cdot)$, and $\bm{\phi}_{\bm{{\bar{\beta}}}}(\cdot)$, respectively. Note that $l_i(\cdot)$ is chosen as $(\bm{f}_i^*-\bm{\hat{f}}_i)^T(\bm{f}_i^* - \bm{\hat{f}}_i)$ in this work for $i = 1, 2, \cdots, N$. For the convenience of subsequent expressions, define $\bm{\varsigma} = \left[\bm{{\bar{\xi}}}^T, \bm{{\bar{\alpha}}}^T, \bm{{\bar{\beta}}}^T\right]^T{\in \mathbb{R}^{n_x(3n-2)}}$, one can achieve
\begin{align} \label{eq_optimal_obj}
	\mathop {\min }\limits_{\bm{a}_f} J(\bm{a}_f) & = \mathop {\min }\limits_{\bm{a}_f} \sum\limits_{i = 1}^{N - 1}l_i(\bm{f}_i^*, \bm{\hat{f}}_i, \bm{a}_f) + l_N(\bm{f}_i^*, \bm{\hat{f}}_i) \notag \\
	s.t.\quad \bm{\varsigma}_{i+1} & = \bm{\phi}^d_{\bm{\varsigma}}(\bm{\varsigma}_i, \bm{a}_f), \quad i = {1,2,\cdots,N-1}
\end{align}
with $\bm{\phi}^d_{\bm{\varsigma}}(\cdot) = \left[\bm{\phi}_{\bm{{\bar{\xi}}}}^d(\cdot)^T,\bm{\phi}_{\bm{{\bar{\alpha}}}}^d(\cdot)^T, \bm{\phi}_{\bm{{\bar{\beta}}}}^d(\cdot)^T \right]^T$. $\bm{\phi}^d_{\bm{\varsigma}}(\bm{\varsigma}_i, \bm{a}_f)$ is further simplified as $\bm{\phi}_{\bm{{\varsigma}},i}^d$ for convenience.

Next, the \textit{Lagrange} multiplier method is utilized to solve Eq. \eqref{eq_optimal_obj}, define \textit{Lagrange} function
\begin{align*}
	\mathcal{L} = J(\bm{a}_f) + \sum\limits_{i = 1}^{N - 1} \bm{\lambda}_{\bm{\varsigma},i}^T (\bm{\phi}_{\bm{{\varsigma}},i}^d-\bm{{\varsigma}}_{i+1})
\end{align*}
with \textit{Lagrange} multiplier $\bm{\lambda}_{\bm{\varsigma},i} {\in \mathbb{R}^{n_x(3n-2)}}$ for $i = 1, 2, \cdots, N-1$. The first-order optimality conditions of the learning problem can be derived as
\begin{align*}
	\frac{{\partial  \mathcal{L}}}{\partial {\bm{\varsigma}}} = \bm{0}, \quad \frac{{\partial  \mathcal{L}}}{\partial \bm{\lambda}_{\bm{{\varsigma}}}} = \bm{0}, \quad \frac{{\partial  \mathcal{L}}}{\partial {\bm{{a}}}} = \bm{0}.
\end{align*}
Thus, one can render
\begin{align}
	\bm{\lambda}_{\bm{{\varsigma}},i-1} &= \nabla_{\bm{\varsigma}_i} l_i + \bm{\lambda}_{\bm{{\varsigma}},i}^T \nabla_{\bm{\varsigma}_i} \bm{\phi}_{\bm{\varsigma},i}^d, \  \bm{\lambda}_{\bm{{\varsigma}},N-1} = \nabla_{\bm{\varsigma}_N} l_N, \label{eq_grad_1}\\
	\bm{{\varsigma}}_{i+1}  &= \bm{\phi}_{\bm{{\varsigma}},i}^d, \quad \bm{\varsigma}_{1} = \bm{\varsigma}(0), \label{eq_grad_2}\\
	\nabla_{\bm{a}_f} \mathcal{L} & = \sum\limits_{i = 1}^{N - 1}(\nabla_{\bm{a}_f} l_i + \bm{\lambda}_{\bm{\varsigma},i}^T \nabla_{\bm{a}_f} \bm{\phi}_{\bm{\varsigma},i}^d) = \bm{0}.  \label{eq_grad_3}
\end{align}

The optimal parameter $\bm{a}_f$ that minimizes $\mathcal{L}$ can be obtained by solving the linear system above. However, directly solving Eqs. \eqref{eq_grad_1}-\eqref{eq_grad_3} becomes computationally expensive, particularly when $N$ and $n_x$ are large. To address this, starting from an initial guess $\bm{a}_{f0}$, we employ an iterative scheme to compute $\bm{a}_f$.

At first, the analytical gradient computation $\nabla_{\bm{a}_f} \mathcal{L}$ is driven by sequentially doing forward rollout Eq. \eqref{eq_grad_2} and backward rollout Eq. \eqref{eq_grad_1}, where the latter one is also known as the term \textit{adjoint solve} or \textit{reverse-mode differentiation}~\citep{chen2018neural}. The calculation process of $\nabla_{\bm{a}_f} \mathcal{L}$ is summarized in Algorithm \ref{gradient_algo}. 

\begin{algorithm}[htb]
	\caption{Analytic gradient computation}
	\label{gradient_algo}
	\begin{algorithmic}[1]
		\Require Learning objective $l_i(\cdot), l_N(\cdot)$; model $\bm{\phi}_{\bm{\varsigma}}^d(\cdot)$; 
		continuous trajectories $\{\bm{{f}}^*(t), \bm{x}(t), \bm{z}(t)\}$.
		\Ensure Gradient $\nabla_{\bm{a}_f} \mathcal{L}$.\\
		\Comment\ Parameter $\bm{a}_{f0}$.
		\State{$\bm{\varsigma}$  $\leftarrow$ Forward rollout of $\bm{\phi}_{\bm{{\varsigma}}}^d(\cdot)$ using Eq. \eqref{eq_grad_2};}
		\State{Compute $\nabla_{\bm{\varsigma}_i} l_i$, $\nabla_{\bm{\varsigma}_i} \bm{\phi}_{\bm{\varsigma},i}^d$, $\nabla_{\bm{\varsigma}_N} l_N$, $\nabla_{\bm{a}_f} l_i$, $\nabla_{\bm{a}_f} \bm{\phi}_{\bm{\varsigma},i}^d$; }
		\State{$\bm{\lambda}_{\bm{\varsigma}}$ $\leftarrow$ Reverse rollout using Eq. \eqref{eq_grad_1};}
		\State{$\nabla_{\bm{a}_f} \mathcal{L}$ $\leftarrow$ Compute gradient using Eq. \eqref{eq_grad_3}.}
		
	\end{algorithmic}
\end{algorithm}

With updated $\nabla_{\bm{a}_f} \mathcal{L}$, the parameter $\bm{a}_f$ is optimized according to the gradient descent algorithm, which is summarized in Algorithm \ref{algo_joint_learning}. Note that the gradient computation is only supported for a single continuous state trajectory, with computational complexity scaling linearly with trajectory length. In practical applications, multiple long-horizon trajectory segments may be employed. In Algorithm \ref{algo_joint_learning}, the mini-batching and stochastic optimization methods are employed to facilitate learning results. The learning step can be set using \textit{Adam} or other stochastic gradient descent-related approaches. An early stopping mechanism is adopted as the convergence condition, which can markedly improve the training efficiency. 

\begin{algorithm}[htb]
	\caption{Training filter parameters}
	\label{algo_joint_learning}
	\begin{algorithmic}[1]
		\Require{Objective $l_k(\cdot), l_N(\cdot)$;
			mini-batch size $s$; trajectories $\{\bm{{f}}^*(t), \bm{x}(t), \bm{z}(t)\}$.}
		\Ensure{Parameter $\bm{a}_f$.} \\
		\Comment\ {Network parameter $\bm{\theta}$; filter parameter $\bm{a}_{f0}$; slice $\mathcal{D}^{tra}$ into $M$ segments $\{\mathcal{D}^{tra}_{j = 1, \cdots, M}\}$ with $s$ length each.}
		\Repeat 
		\For{$\{\bm{{f}}^*_{1:s}, \bm{x}_{1:s}, \bm{z}_{1:s}\}$ in $\{\mathcal{D}^{tra}_{j = 1, \cdots, M}\}$}
		\State{Compute analytic gradient $\nabla_{\bm{a}_f} \mathcal{L}$ using Algorithm \ref{gradient_algo};}
		\State{Compute learning step $\eta \leftarrow Optimizer(\bm{a}_f, \nabla_{\bm{a}_f} \mathcal{L})$;}
		\State{$\bm{a}_f \leftarrow \bm{a}_f - \eta$.}
		\EndFor
		\Until{\textbf{convergence}}
	\end{algorithmic}
\end{algorithm}

The learning performance of Algorithm \ref{algo_joint_learning} is verified in the acceleration example, as shown in Fig. 6e. More experimental results are provided in Supplementary Fig. S2.

\section{Filtering learning residual with more general form} \label{sec_general_filter}
In comparison with Eq. (\textcolor{blue}{3}), a more general filter is considered in this part, i.e.,
\begin{align} \label{eq_n_filter_gener}
	\mathcal{L} \left[\mathcal{F}(\bm{\gamma}(t))\right] = \frac{b_{n-1}s^{n-1} + \cdots +b_{1}s + b_0}{s^n + a_{n-1}s^{n-1} + \cdots +a_{1}s + a_0}{\Upsilon}(s)
\end{align}
with more parameters $b_{n-1}, \cdots, b_{1}, b_{0}$. The above filter can cover the bandpass case. By substituting Eq. \eqref{eq_n_filter_gener} into Eq. (\textcolor{blue}{1}) and performing the result in time-domain, it can be rendered that 
\begin{align}
	& \bm{\hat{f}}^{(n)}(t) + a_{n-1}\bm{\hat{f}}^{(n-1)}(t) + \cdots +a_1 \bm{\hat{f}}{'}(t) + a_0 \bm{\hat{f}}(t) \notag\\ = & \bm{f}_{\bm{\theta}}^{(n)}(t) + a_{n-1}\bm{f}_{\bm{\theta}}^{(n-1)}(t) + \cdots +a_1 \bm{f}_{\bm{\theta}}'(t) + a_0 \bm{f}_{\bm{\theta}}(t) \notag\\ & + b_{n-1}\bm{\gamma}^{n-1}(t) + \cdots +b_1 \bm{\gamma}'(t) + b_0 \bm{\gamma}(t).
\end{align}

Define $\bm{\chi}_1(t) = {a_{n - 1}}\int {\bm {\hat{f}}(t)dt}  +  \cdots +{a_1}{\int}^{n - 1}\bm{\hat {f}}(t)(dt)^{n - 1}$, $\bm{\chi}_2(t) = \bm{f}_{\bm{\theta}}(t) + ({a_{n - 1}}-{b_{n - 1}})\int {\bm {{f}}_{\bm{\theta}}(t)dt}  +  \cdots +({a_1}-b_1){\int}^{n - 1}\bm{{f}}_{\bm{\theta}}(t)(dt)^{n - 1}$, $\bm{\chi}_3(t) =  b_{n-1}\int \bm{g}(t)dt + \cdots + b_{n-1}\int^{n - 1}\bm{g}(t)(dt)^{n - 1}$, and recall $\bm{\gamma}(t) = \bm{\dot{x}} - \bm{g}(t) - \bm{{f}}_{\bm{\theta}}(t)$. One can obtain
\begin{align}
	& \bm{\hat f}^{(n)}(t) + \bm{\chi}_1^{(n)} + a_0 \bm{\hat{f}}(t) \notag \\ = & \bm{\chi}_2^{(n)} - \bm{\chi}_3^{(n)}+ a_0 \bm{f}_{\bm{\theta}}(t) - b_0\bm{g}(t) -b_0\bm{f}_{\bm{\theta}}(t) + b_{n-1}\bm{x}^{(n)}+\cdots +b_0\bm{\dot{x}}.
\end{align}
Define an auxiliary variable $\bm{\xi} = \bm{\hat f}(t) + \bm{\chi}_1- \bm{\chi}_2 + \bm{\chi}_3 - b_{n-1}\bm{x}- \cdots -b_0 {\int}^{n-1}\bm{x}(dt)^{n - 1}$. It can be rendered that
\begin{align} \label{eq_observer_general}
	\left\{ {\begin{array}{*{20}{c}}
			\bm{\xi}^{(n)} = (a_0 -b_0)\bm{f}_{\bm{\theta}}(t) -  a_0\bm{g}(t) - a_0 \bm{\hat{f}}(t)\\
			\bm{\hat{f}}(t) = \bm{\xi} - \bm{\chi}_1 + \bm{\chi}_2 -\bm{\chi}_3 + b_{n-1}\bm{x} +\cdots +b_0 {\int}^{n-1}\bm{x}(dt)^{n - 1}.
	\end{array}} \right.
\end{align}
Compared with Eq. (\textcolor{blue}{5}), Eq. \eqref{eq_observer_general} can capture more complex leanring residual at the cost of complex parameter adjustment.

\section{Several special forms} \label{sec_special_case}
In this part, we will show the feedback neural network \citep{jia2024feedback},  EVOLVER \citep{10288520}, and SEER-II \citep{jia2025} belong to the framework of PhyFilter. Consider the first-order form of the PhyFilter,
\begin{align} \label{eq_observer_1order}
	\left\{ {\begin{array}{*{20}{c}}
			\bm{\dot{\xi}} = -  \bm{g}(t) - \bm{\hat{f}}(t)\\
			\bm{\hat{f}}(t) = a_0\bm{\xi} + \bm{f}_{\bm{\theta}}(t) + a_0 \bm{x}.
	\end{array}} \right.
\end{align}

\subsection{Feedback neural network}
Substitute the first subequation into the second one of Eq. \eqref{eq_observer_1order} can yield
\begin{align}
	\bm{\hat{f}}(t) = - a_0 \int_{{t_0}}^t {(\bm{g}(t) + \bm{\hat{f}}(t))dt} + \bm{f}_{\bm{\theta}}(t) + a_0 \bm{x}.
\end{align}
Define a state estimation $\bm{\hat x} = \int_{{t_0}}^t {(\bm{g}(t) + \bm{\hat{f}}(t))}dt$ as in \citep{jia2024feedback}, one can obtain
\begin{align}
	\bm{\hat{f}}(t) = \bm{f}_{\bm{\theta}}(t) + a_0 (\bm{x}- \bm{\hat x}).
\end{align}
The above equation is exactly the key feedback equation \cite[Eq. (7)]{jia2024feedback}, where $\bm{f}_{\bm{\theta}}(t)$ is learned through the neural ODE by providing state trajectories and $a_0$ denotes the feedback gain.

\subsection{\textit{Koopman} operator-based uncertainty observer}
By defining $\bm{\xi}_1  = {a_0}\bm{\xi}  + \bm{f}_{\bm{\theta}}(t)$, Eq. \eqref{eq_observer_1order} can be adjusted to
\begin{align} \label{eq_evolver}
	\left\{ {\begin{array}{*{20}{c}}
			\bm{\dot{\xi}}_1 = \bm{\dot{f}}_{\bm{\theta}}(t) - a_0(\bm{g}(t)+\bm{\hat{f}}(t))\\
			\bm{\hat{f}}(t) = \bm{\xi}_1 + a_0 \bm{x}.
	\end{array}} \right.
\end{align}
By regarding $- a_0$ and $\bm{{f}}_{\bm{\theta}}(t)$ as the observer gain and unknown uncertainty, respectively, the above equation is exactly the uncertainty observer \cite[Eq. (19)]{10288520}. Note that $\bm{{f}}_{\bm{\theta}}(t)$ is learned through the online \textit{Koopman} operator in \citep{10288520}, which involves an online construction process of the training dataset. Theoretically, the EVOLVER \citep{10288520} is devised targeted at the uncertainty with $\bm{\dot{f}}(t) = \bm{h}(\bm{{f}}(t), \bm{x}, \bm{d})$ form, where $\bm{d}$ denotes external factors.

\subsection{\textit{Chebyshev}-based uncertainty observer}
The basic idea of the SEER-II in \citep{jia2025} is similar to the EVOLVER \citep{10288520}, apart from the term $\bm{\dot{f}}_{\bm{\theta}}(t)$ is handled by \textit{Chebyshev} variable decomposition. The targeted uncertainty is modeled as $\bm{{f}}(t) = \bm{h}(\bm{x}, \bm{d})$. With the analytic assumption, it can be proven $\bm{{f}}(t)$ can be decomposed into $\bm{\Xi}_d \bm{\Phi}(t)\bm{\varsigma}(\bm{x})$ with arbitrarily high precision. $\bm{\Xi}_d$ is a learned parameter matrix. $\bm{\Phi}(t)$ and $\bm{\varsigma}(\bm{x})$ are composed of \textit{Chebyshev} polynomials. Hence, $\bm{\dot{f}}_{\bm{\theta}}(t)$ can be analytically represented as
\begin{align}
	\bm{\dot{f}}_{\bm{\theta}}(t) = \bm{\Xi}_d \bm{\dot{\Phi}}(t)\bm{\varsigma}(\bm{x}) +  \bm{\Xi}_d \bm{{\Phi}}(t)\frac{{\partial \bm{\varsigma}(\bm{x})}}{{\partial \bm{x}}}\bm{\dot{x}}.
\end{align}
By defining $\bm{\xi}_2  = \bm{\xi}_1 - \int {\bm{\Xi}_d \bm{{\Phi}}(t)\frac{{\partial \bm{\varsigma}(\bm{x})}}{{\partial \bm{x}}} d\bm{x}}$, Eq. \eqref{eq_evolver} can be adjusted to
\begin{align} \label{eq_seerii}
	\left\{ {\begin{array}{*{20}{c}}
			\bm{\dot{\xi}}_2 =  \bm{\Xi}_d \bm{\dot{\Phi}}(t)\bm{\varsigma}(\bm{x}) - a_0(\bm{g}(t)+\bm{\hat{f}}(t))\\
			\bm{\hat{f}}(t) = \bm{\xi}_2 + \int {\bm{\Xi}_d \bm{{\Phi}}(t)\frac{{\partial \bm{\varsigma}(\bm{x})}}{{\partial \bm{x}}} d\bm{x}} + a_0 \bm{x}
	\end{array}} \right.
\end{align}
which is exactly the SEER-II \cite[Eq. (11)]{jia2025} with observer gain $a_0$.

Compared with Eq. \eqref{eq_observer_1order}, EVOLVER \eqref{eq_evolver} and SEER-II \eqref{eq_seerii} move differentiable learned uncertainty to the first ODE subequation. By this way, the possible discontinuous problem induced by learning can be alleviated to a certain extent.

\section{Joint learning}
A learning algorithm is presented in this part to learn the filter parameter $\bm{a}_f = \{a_0, a_1, \cdots, a_{n-1}\}$ in Eq. (\textcolor{blue}{3}) with model parameter $\bm{\theta}$ together.  

Following the definition in Section ``Learning filter parameter'', define $\bm{\omega} = \{\bm{a}_f, \bm{\theta}\}$. Regard the equations Eqs. (\textcolor{blue}{6})-(\textcolor{blue}{7}) as an optimal control system with constrained states $\{\bm{{\bar{\xi}}}, \bm{{\bar{\alpha}}}, \bm{{\bar{\beta}}}\}$ and control input $\bm{\omega}$. Thus, the optimization problem to find optimal $\bm{\omega}$ can be formalized as 
\begin{align} \label{eq_optimal_obj2}
	\mathop {\min }\limits_{\bm{\omega}} J(\bm{\omega}) & = \mathop {\min }\limits_{\bm{\omega}} \sum\limits_{i = 1}^{N - 1}l_i(\bm{f}_i^*, \bm{\hat{f}}_i, \bm{\omega}) + l_N(\bm{f}_i^*, \bm{\hat{f}}_i) \notag \\
	s.t.\quad \bm{\varsigma}_{i+1} & = \bm{\phi}^d_{\bm{\varsigma}}(\bm{\varsigma}_i, \bm{\omega}), \quad i = {1,2,\cdots,N-1}.
\end{align}
The \textit{Lagrange} multiplier method is utilized to slove Eq. \eqref{eq_optimal_obj2}, define \textit{Lagrange} function
\begin{align}
	\mathcal{L} = J(\bm{\omega}) + \sum\limits_{i = 1}^{N - 1} \bm{\lambda}_{\bm{\varsigma},i}^T (\bm{\phi}_{\bm{{\varsigma}},i}^d-\bm{{\varsigma}}_{i+1}).
\end{align}
The solving procedure for optimal $\bm{\omega}$ is similar to $\bm{a}_f$ in Section ``Learning filter parameter''. Especially, $\nabla_{\bm{\omega}} l_i$ and $\nabla_{\bm{\omega}} \bm{\phi}_{\bm{\varsigma},i}^d$ need to be derived analytically.

\section{Trajectory generation of aerial manipulator} \label{sec_traj}
Focusing on achieving the aerial interaction mission, we formulate an optimization problem to generate a $6$-dimensional motion state, which includes the aerial platform’s trajectory in the inertia frame and the end-effector's trajectory in its base frame. Firstly, the state propagation of the aerial manipulator is given. Let $h=\left[P_b^{\top},{\dot{P}}_b^{\top},P_{ed}^{\top},{\dot{P}}_{ed}^{\top}\right]^{\top}\in\mathbb{R}^{12\times1}$ be the controlled state vector. The propagation of the controlled state $h_i$ in discrete time can be formulated as
\begin{equation}
	\left\{\begin{matrix}h_{i+1}=\textbf{{\textit{A}}}h_i+\textbf{{\textit{B}}}u_i\\y_{i+1}=\textbf{{\textit{C}}}h_{i+1}\\\end{matrix}\right.
\end{equation}
where $u_i=\left[{\ddot{P}}_b^{\top},{\ddot{P}}_{ed}^{\top}\right]_i^{\top}\in\mathbb{R}^{6\times1}$ is the optimized control action consisting of the accelerations of the aerial platform in the inertia frame and the end-effector with respect to its base. Furthermore, $y_{i+1}=\left[y^{pu^{\top}},y^{vu^{\top}},y^{pm^{\top}},y^{vm^{\top}}\right]_{i+1}^{\top}\in\mathbb{R}^{12\times1}$ represents the predictive model output. The tracking error of the end-effector can be formulated as
\begin{subequations}
	\begin{align}
		e_{P}(t+i)&=P_e^d(t+i)-{\hat{P}}_e(t+i)\\
		{\hat{P}}_e(t+i)&=y^{pu}(t+i)-\textbf{\textit{{R}}}_b{(t+i)P}_o-\textbf{\textit{{R}}}_b(t+i)y^{pm}(t+i)\\
		\textbf{\textit{{R}}}_b(t+i)&=\textbf{\textit{{R}}}_b(t+i-1)+\delta t\textbf{\textit{{R}}}_b(t+i-1)\omega_b^\times(t)
	\end{align}
\end{subequations}
where $\delta t$ is the state propagation period. $P_e^d\in\mathbb{R}^{3\times1}$ represents the desired trajectory of the end-effector in the inertia frame. $P_{o}\in\mathbb{R}^{3\times1}$ is the constant deviation between the manipulator base and the CoM of the aerial platform. $\omega_{b}\in\mathbb{R}^{3\times1}$ denotes the angular velocity of the aerial platform.  Therefore, the cooperative planning problem of the aerial manipulator is converted to seek an optimal control input $u$ to minimize the following error cost function
\begin{equation}
	Q_1=\Vert e_{P}\Vert^{2}_{\bm{\Lambda_1}}=e_{P}^{\top}\bm{\Lambda_1}e_{P},    e_{P}\in \mathbb{R}^{3N\times1} \label{Q1V1}
\end{equation}
where $\bm{\Lambda}_1\in\mathbb{R}^{3N\times3N}$ is a positive-definite weighting matrix. If the optimization formulation 
has no constraints or does not trigger any constraints, Eq. (\ref{Q1V1}) can be simplified to an unconstrained quadratic programming (QP) problem. Nevertheless, such a pure formulation neglects practical safety during executing dynamic operations and fails to account for the motion characteristics of the aerial manipulator. Therefore, specific constraints are of seminal importance that must be incorporated in the planner design phase.

\noindent{{\textbf{Control action smoothness:}}} The smoothness of control action $u$ is a major step to prevent unnecessary aggressive control responses, thereby enhancing system safety. 
For such a safe concern, the following cost function for optimizing the control input $u$ is considered as
\begin{equation}
	Q_2=\Vert \Delta u\Vert^{2}_{\bm{\Lambda_2}}=\Delta u^{\top}\bm{\Lambda_2}\Delta u,    \Delta u\in \mathbb{R}^{6N\times1} \label{Q3}
\end{equation}
where $\bm{\Lambda}_2\in\mathbb{R}^{6N\times6N}$ is a weighting matrix and $\Delta u$ represents the increment of control action at each step. The additional cost can mitigate sharp acceleration maneuvers caused by abrupt external impulses acting on the aerial manipulator.

\noindent{{\textbf{State stability:}}} Although the above two-term formulation is widely used in trajectory generation, the neglected state increments may lead to poor output stability of aerial manipulators. It is feasible to control either the aerial platform or the manipulator while maintaining end-effector tracking accuracy~\citep{10339889}. An undesirable scenario may arise wherein both the aerial platform and the manipulator exhibit persistent oscillatory behavior in an attempt to maintain stability of the end-effector. Therefore, a constraint on state increments is imposed to improve the stability of the optimized cooperative trajectory, which is expressed as
\begin{subequations}
	\begin{align}
		&Q_{3,u} = \Vert  \Delta {y}^{pu} \Vert^{2}_{\bm{\Lambda_{3}^{u}}}={\Delta {y}^{pu}}^{\top}\bm{\Lambda_{3}^{u}}\Delta {y}^{pu},    \Delta {y}^{pu}\in \mathbb{R}^{3N\times1}\\
		&Q_{3,m} = \Vert \Delta {y}^{pm} \Vert^{2}_{\bm{\Lambda_{3}^{m}}}={\Delta {y}^{pm}}^{\top}\bm{\Lambda_{3}^{m}}\Delta {y}^{pm},    \Delta {y}^{pm}\in \mathbb{R}^{3N\times1}\\
		&Q_3 = Q_{3,u} + Q_{3,m}
	\end{align}
\end{subequations}
where $\Delta {y}^{pu}$ and $\Delta {y}^{pm}$ represent the predicted output increments of the aerial platform and the end-effector from sample time $t+1$ to $t+N$, respectively. Moreover,  $\bm{\Lambda_{3}^{u}}\in\mathbb{R}^{3N\times3N}$ and $\bm{\Lambda_{3}^{m}}\in\mathbb{R}^{3N\times3N}$ denote the corresponding positive weighting matrices based on their motion characteristics. The inclusion of state increments significantly contributes to increasing the system safety.

The propagation matrices are formulated as
\begin{equation}
	\textbf{\textit{A}}=\left[\begin{matrix}\textbf{\textit{I}}_3&\delta t\textbf{\textit{I}}_3&\textbf{\textit{0}}_\textbf{3}&\textbf{\textit{0}}_3\\\textbf{\textit{0}}_\textbf{3}&\textbf{\textit{I}}_3&\textbf{\textit{0}}_3&\textbf{\textit{0}}_3\\\textbf{\textit{0}}_\textbf{3}&\textbf{\textit{0}}_\textbf{3}&\textbf{\textit{I}}_3&\delta t\textbf{\textit{I}}_3\\
		\textbf{\textit{0}}_\textbf{3}&\textbf{\textit{0}}_\textbf{3}&\textbf{\textit{0}}_3&\textbf{\textit{0}}_3\\\end{matrix}\right]
	\textbf{\textit{B}}=\left[\begin{matrix}0.5\delta t^{2}\textbf{\textit{I}}_3&\textbf{\textit{0}}_3\\\delta t\textbf{\textit{I}}_3&\textbf{\textit{0}}_3\\
		\textbf{\textit{0}}_\textbf{3}&0.5\delta t^{2}\textbf{\textit{I}}_3\\\textbf{\textit{0}}_\textbf{3}&\delta t\textbf{\textit{I}}_3\\\end{matrix}\right]
	\textbf{\textit{C}}=\left[\begin{matrix}\textbf{\textit{I}}_3&\textbf{\textit{0}}_3&\textbf{\textit{0}}_\textbf{3}&\textbf{\textit{0}}_3\\\textbf{\textit{0}}_\textbf{3}&\textbf{\textit{I}}_3&\textbf{\textit{0}}_3&\textbf{\textit{0}}_3\\\textbf{\textit{0}}_\textbf{3}&\textbf{\textit{0}}_\textbf{3}&\textbf{\textit{I}}_3&\textbf{\textit{0}}_3\\\textbf{\textit{0}}_\textbf{3}&\textbf{\textit{0}}_\textbf{3}&\textbf{\textit{0}}_3&\textbf{\textit{I}}_3\\\end{matrix}\right]				
\end{equation}
where $\delta t=0.07$. Moreover, the optimization parameters of the presented cooperative planner are:
$\bm{\Lambda_{1}}$ = diag\{6000, 6000, 9000\}, $\bm{\Lambda_{2}}$ = diag\{700, 700, 700, 1, 1, 1\}, $\bm{\Lambda_{3}^{u}}$ = diag\{20, 20, 20\}, $\bm{\Lambda_{3}^{m}}$ = diag\{10, 10, 10\}.

\section{\textcolor{black}{Robustness analysis under inaccurate priors}}
\label{app:robustness}

\textcolor{black}{This appendix analyzes the behavior of PhyFilter when the physical prior used in the filter is inaccurate, such as in the locomotion case, and derives an explicit bound on the resulting estimation error for a general $n$-th order filter.}

\subsection{\textcolor{black}{Problem setup}}

Recall the physical structure used by PhyFilter in the locomotion case,
\begin{equation}
	\bm{M}(t)\dot{\bm{x}} = \bm{g}(t) + \bm{f}(t),
	\label{eq_app_structure}
\end{equation}
where $\bm{x}=\dot{\bm{q}}$, $\bm{g}(t)$ collects the known terms, and
$\bm{f}(t)=\Delta\bm{\tau}$ is the unknown term to be estimated. In practice
the true inertia matrix $\bm{M}(t)$ is replaced by a simplified
$\hat{\bm{M}}(t)$ (e.g., the diagonalized inertia and linearized \textit{Coriolis}
terms). Define the prior error
\begin{equation}
	\Delta\bm{M}(t) \triangleq \bm{M}(t) - \hat{\bm{M}}(t).
	\label{eq_app_deltaM}
\end{equation}

The $n$-th order PhyFilter reconstructs $\bm{f}(t)$ from the measurements
$\bm{g}(t)$ and $\hat{\bm{M}}(t)$ while enforcing the designed low-pass
dynamics of Eqs.~(3)--(5). Note that the filter is implemented with the
\emph{simplified} prior $\hat{\bm{M}}$, whereas the plant~\eqref{eq_app_structure}
evolves with the true $\bm{M}$. Following the same substitution procedure that
leads from Eq.~(3) to Eq.~(5), and using $\bm{g}=\bm{M}\dot{\bm{x}}-\bm{f}$ from
the true plant Eq.~\eqref{eq_app_structure}, replacing $\bm{M}$ by $\hat{\bm{M}}$
introduces the additional forcing $-a_0\Delta\bm{M}\dot{\bm{x}}$ on the
right-hand side while the left-hand side retains the designed filter structure
\begin{equation}
	\hat{\bm{f}}^{(n)} + a_{n-1}\hat{\bm{f}}^{(n-1)} + \cdots + a_1\dot{\hat{\bm{f}}}
	+ a_0\hat{\bm{f}} = a_0\big(\bm{f} - \Delta\bm{M}\,\dot{\bm{x}}\big).
	\label{eq_app_highorder_io}
\end{equation}

Define the estimation error $\tilde{\bm{f}} \triangleq \bm{f} - \hat{\bm{f}}$.
Subtracting Eq.~\eqref{eq_app_highorder_io} from the corresponding derivatives of
$\bm{f}$, the $a_0\bm{f}$ term cancels and each component of the estimation error
obeys
\begin{equation}
	\tilde{\bm{f}}^{(n)} + a_{n-1}\tilde{\bm{f}}^{(n-1)} + \cdots
	+ a_1\dot{\tilde{\bm{f}}} + a_0\tilde{\bm{f}} = \bm{d}(t),
	\label{eq_app_err_lhs}
\end{equation}
\begin{equation}
	\bm{d}(t) \triangleq \underbrace{\bm{f}^{(n)} + a_{n-1}\bm{f}^{(n-1)} + \cdots + a_1\dot{\bm{f}}}_{\text{intrinsic residual dynamics}} + a_0\,\Delta\bm{M}\,\dot{\bm{x}}.
	\label{eq_app_highorder_err}
\end{equation}
Eq.~\eqref{eq_app_highorder_err} is the central relation: the prior error
$\Delta\bm{M}$ enters the estimator dynamics \emph{only} through the forcing
$\bm{d}(t)$, alongside the intrinsic residual dynamics, and does
not alter the characteristic polynomial $s^n + a_{n-1}s^{n-1} + \cdots + a_0$ of
the filter.

\subsection{\textcolor{black}{Boundedness analysis}}

Introducing the augmented state
$\bm{z} \triangleq [\,\tilde{\bm{f}}^\top,\dot{\tilde{\bm{f}}}^\top,\ldots,
(\tilde{\bm{f}}^{(n-1)})^\top\,]^\top$, Eq.~\eqref{eq_app_highorder_err} is written
in companion form
\begin{equation}
	\dot{\bm{z}} = \bm{A}\,\bm{z} + \bm{B}\,\bm{d}(t),
	\quad
	\bm{A} =
	\begin{bmatrix}
		\bm{0} & \bm{I} & & \\
		& \ddots & \ddots & \\
		& & \bm{0} & \bm{I} \\
		-a_0\bm{I} & -a_1\bm{I} & \cdots & -a_{n-1}\bm{I}
	\end{bmatrix},
	\quad
	\bm{B} =
	\begin{bmatrix}
		\bm{0}\\ \vdots \\ \bm{0} \\ \bm{I}
	\end{bmatrix}.
	\label{eq_app_ss}
\end{equation}
Since the filter coefficients are placed, either by pole placement or by the
auto-learning algorithm, such that $s^n + a_{n-1}s^{n-1} + \cdots + a_0$ is
Hurwitz, $\bm{A}$ is Hurwitz, and for any $\bm{Q}=\bm{Q}^\top\succ 0$ there
exists $\bm{P}=\bm{P}^\top\succ 0$ satisfying
$\bm{A}^\top\bm{P}+\bm{P}\bm{A}=-\bm{Q}$.

Suppose the residual derivatives and the prior error are bounded, i.e.,
$\|\bm{f}^{(k)}(t)\|\le \delta_f$ for $k=1,\ldots,n$,
$\|\Delta\bm{M}(t)\|\le \delta_M$, and $\|\dot{\bm{x}}(t)\|\le \delta_x$ for all
$t$, with $\delta_f,\delta_M,\delta_x \ge 0$. These conditions merely state that
the lumped residual varies at a bounded rate and that the modeling error and
operating velocity are bounded, all of which hold in practice. Then the intrinsic
part of the forcing satisfies
$\|\bm{f}^{(n)} + a_{n-1}\bm{f}^{(n-1)} + \cdots + a_1\dot{\bm{f}}\|
\le c_f\,\delta_f$ with $c_f \triangleq 1 + \sum_{i=1}^{n-1} a_i$, and hence
\begin{equation}
	\|\bm{d}(t)\|\le c_f\,\delta_f + a_0\,\delta_M\,\delta_x \triangleq \bar{d}.
	\label{eq_app_dbar}
\end{equation}
Taking $V(\bm{z}) = \bm{z}^\top\bm{P}\bm{z}$ and differentiating
along~Eq.~\eqref{eq_app_ss},
\begin{equation}
	\dot V
	= -\bm{z}^\top\bm{Q}\bm{z} + 2\,\bm{z}^\top\bm{P}\bm{B}\,\bm{d}
	\le -\lambda_{\min}(\bm{Q})\,\|\bm{z}\|^2
	+ 2\,\lambda_{\max}(\bm{P})\,\|\bm{B}\|\,\|\bm{z}\|\,\bar{d}.
	\label{eq_app_vdot}
\end{equation}
Hence $\dot V<0$ whenever
$\|\bm{z}\| > 2\lambda_{\max}(\bm{P})\|\bm{B}\|\,\bar{d}/\lambda_{\min}(\bm{Q})$,
so $\bm{z}$, and therefore the estimation error $\tilde{\bm{f}}$, is uniformly
ultimately bounded, with an ultimate bound proportional to
\begin{equation}
	\limsup_{t\to\infty}\|\tilde{\bm{f}}(t)\|
	\;\propto\;
	\bar{d}
	=
	\underbrace{c_f\,\delta_f}_{\text{intrinsic}}
	+ \underbrace{a_0\,\delta_M\,\delta_x}_{\text{prior-induced}}.
	\label{eq_app_bound}
\end{equation}

Three observations follow. First, for any bounded prior error
$\Delta\bm{M}$, the error dynamics remain governed by the Hurwitz matrix
$\bm{A}$. An inaccurate prior only enlarges the forcing $\bm{d}(t)$ and thus the
ultimate bound, but never breaks stability. This is why the diagonalized
$\bm{M}$ and linearized $\bm{C}$ adopted in the locomotion case do not
compromise performance. Second, the bound decomposes into an intrinsic part,
scaling with the residual variation rate $\delta_f$ and present even under a
perfect prior, and a prior-induced part, scaling with the modeling error
$\delta_M$ and the operating velocity $\delta_x$. The latter vanishes when the
prior is exact ($\delta_M=0$). Third, the prior-induced disturbance
$a_0\Delta\bm{M}\dot{\bm{x}}$ enters through the same channel $\bm{B}$ as any
other residual and is attenuated by the Hurwitz error dynamics, so the modeling
error is treated as one more component of the lumped residual to be suppressed
rather than a source of divergence, which is precisely why PhyFilter tolerates
a simplified, or even substantially inaccurate, physical prior.

\section{Experimental hardware} \label{sec_hardware}

In this section, the hardware components utilized in experiments, including a quadruped robot, an aerial drone, and an aerial manipulator, are introduced.

For the locomotion experiment, the DEEPRobotics Lite3 quadruped robot is employed, which comprises a locomotion host (equipped with an RK3588 CPU) and a perception host (equipped with an NVIDIA Jetson Xavier NX). The robot has a weight of $11$ kg and dimensions of $548 \times 370 \times 118$ mm$^3$. During employment, the locomotion control policy is executed on a remote personal computer, with control commands transmitted to the robot via a Wi-Fi connection. 

In the experiments involving drone flight and aerial manipulation, the same coaxial dual-rotor octocopter platform is utilized. For state perception, an STM32F427 microprocessor is employed to receive data from both the motion capture system and the onboard IMU, while another STM32F427 microprocessor is dedicated to low-level control. A UWB (Ultra-Wideband) communication module is employed for signal transmission between the drone and the ground station. To enable online planning of pick-and-place trajectories, an Intel NUC is integrated as the onboard computing unit for the aerial manipulator. Additionally, a $5$-DOF manipulator is mounted on the drone to execute aerial manipulation tasks.

\section{Quadruped tests under full domain randomization} \label{rl_full}

\textcolor{black}{Fig. 2 shows the training setup using only flat terrain. To further strengthen the comparison, we additionally evaluate a stronger setting in which both policies are trained on full terrains and tested on more challenging ones. Specifically, both the baseline and the PhyFilter-augmented policy are trained with standard domain randomization, including diverse terrains (curriculum: 0–0.90), payload variations, and disturbances, rather than the simplified flat-terrain setting. We evaluate them under a more challenging, out-of-distribution scenario (curriculum: 0.92–1.00) that lies beyond the training distribution. As shown in Fig. \ref{full_dr}, even when the baseline is trained with full standard randomization, it still fails under sufficiently OOD conditions, whereas PhyFilter continues to provide additional benefit on top of it. Moreover, we deployed this fully trained baseline on the real-world terrains in Fig. 3, and found its performance inferior to that of PhyFilter, even though the latter was trained purely on flat terrain (see Supplementary Movie 6 for details).}

\section{Humanoid tests}
\textcolor{black}{To further verify effectiveness, we added a humanoid experiment in this revision. Notably, to mitigate the modeling uncertainty considered by the reviewer, the feedforward model here is computed via a state-synchronized parallel MuJoCo simulation. The humanoid policy is trained with standard randomization ranges. At test time, we apply both in-distribution (ID) and out-of-distribution (OOD) external forces to one of its arms. As shown in Fig. \ref{appen_humanoid}, PhyFilter keeps the robot stably standing under both conditions, whereas the baseline becomes unstable under the OOD force.}

\end{document}